\documentclass[letterpaper]{article} % DO NOT CHANGE THIS
\usepackage{aaai25}  % DO NOT CHANGE THIS
\usepackage{times}  % DO NOT CHANGE THIS
\usepackage{helvet}  % DO NOT CHANGE THIS
\usepackage{courier}  % DO NOT CHANGE THIS
\usepackage[hyphens]{url}  % DO NOT CHANGE THIS
\usepackage{graphicx} % DO NOT CHANGE THIS
\usepackage{natbib}  % DO NOT CHANGE THIS AND DO NOT ADD ANY OPTIONS TO IT
\usepackage{caption} % DO NOT CHANGE THIS AND DO NOT ADD ANY OPTIONS TO IT
\usepackage{amssymb}
\usepackage{mhchem}
\usepackage[ruled,vlined]{algorithm2e}

\usepackage{amsmath}
\usepackage{adjustbox}
\usepackage{multirow}
\usepackage{subcaption}
\usepackage{anyfontsize}
\usepackage{blindtext}
\usepackage{multirow}
\usepackage{pifont}% http://ctan.org/pkg/pifont
\usepackage{booktabs} % For formal tables
\usepackage{color, colortbl}
\usepackage[dvipsnames]{xcolor}
\usepackage{tikz}
\usepackage{extarrows}
\newcommand{\dashleftrightarrow}{\dashleftarrow\dashrightarrow}

\usepackage{float}

\usepackage{newfloat}
\usepackage{listings}
\DeclareCaptionStyle{ruled}{labelfont=normalfont,labelsep=colon,strut=off} % DO NOT CHANGE THIS
\floatstyle{ruled}
\newfloat{listing}{tb}{lst}{}
\floatname{listing}{Listing}
\title{Context-aware Graph Causality Inference for Few-Shot Molecular Property Prediction}

\author{
    Van Thuy Hoang and O-Joun Lee\thanks{Corresponding author: O-Joun Lee (Tel.: +82-2-2164-5516)}
}
\affiliations{
    Department of Artificial Intelligence, The Catholic University of Korea \\
    {\{hoangvanthuy90, ojlee\}@catholic.ac.kr}
}

\usepackage{bibentry}
\begin{document}

\maketitle

% \begin{abstract}
% AAAI creates proceedings, working notes, and technical reports directly from electronic source furnished by the authors. To ensure that all papers in the publication have a uniform appearance, authors must adhere to the following instructions.
% \end{abstract}
\begin{abstract}

Molecular property prediction is becoming one of the major applications of graph learning in Web-based services, e.g., online protein structure prediction and drug discovery.
A key challenge arises in few-shot scenarios, where only a few labeled molecules are available for predicting unseen properties.
Recently, several studies have used in-context learning to capture relationships among molecules and properties, but they face two limitations in:
(1) exploiting prior knowledge of functional groups that are causally linked to properties and
(2) identifying key substructures directly correlated with properties.
We propose CaMol, a context-aware graph causality inference framework, to address these challenges by using a causal inference perspective, assuming that each molecule consists of a latent causal structure that determines a specific property.
First, we introduce a context graph that encodes chemical knowledge by linking functional groups, molecules, and properties to guide the discovery of causal substructures.
Second, we propose a learnable atom masking strategy to disentangle causal substructures from confounding ones.
Third, we introduce a distribution intervener that applies backdoor adjustment by combining causal substructures with chemically grounded confounders, disentangling causal effects from real-world chemical variations.
Experiments on diverse molecular datasets showed that CaMol achieved superior accuracy and sample efficiency in few-shot tasks, showing its generalizability to unseen properties.
Also, the discovered causal substructures were strongly aligned with chemical knowledge about functional groups, supporting the model interpretability.

\end{abstract}

\section{Introduction}

% \listofappendixtables
% \listofappendixfigures

% \renewcommand{\thetable}{A\arabic{table}}
% \renewcommand{\thefigure}{A\arabic{figure}}
% \renewcommand{\theequation}{A\arabic{equation}}

%The goal is to predict the properties of unlabeled molecules by using a few numbers of labeled molecules, which enables generalization to novel tasks with minimal supervision.
\begin{figure}[t]
\centering
  \includegraphics[width=  \linewidth]{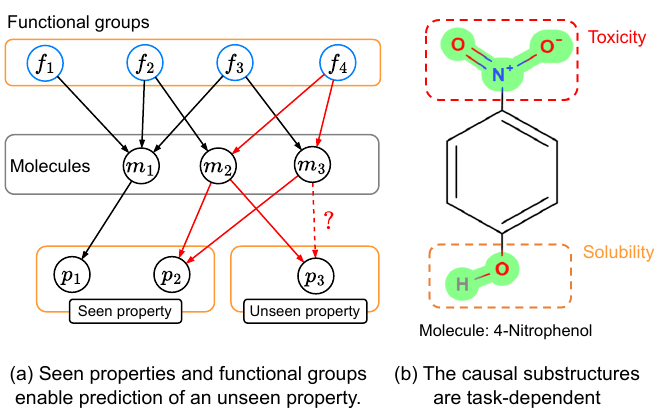}
  \caption{ 
  (a) The seen properties are relevant to the unseen property prediction.
  %The causal substructure, Hydroxyl group (\ce{OH}), is the main reason to determine the solubility, while the remaining substructures are irrelevant.
  (b) The causal substructures vary and depend on molecular property prediction tasks.
  }
  \label{fig:problem}
\end{figure}

Molecular Property Prediction (MPP) is increasingly being integrated into Web-based platforms, e.g., AlphaFold\footnote{\url{https://alphafold.ebi.ac.uk}} \cite{jumper2021highly}, which provide protein structure prediction online, and drug discovery platforms, e.g., DrugFlow\footnote{\url{https://chem-space.com}}, which recommend candidate molecules for therapeutic properties \cite{DBLP:journals/jcisd/ShenSHCKYWWZZZCCCLZJCJW24,DBLP:journals/jcisd/RusinkoRFBKG24}.
In this setting, molecules are represented as graphs, and the task is to predict property labels such as toxicity or solubility from their structural features
However, acquiring labeled molecules remains a challenge, as wet-lab experiments commonly used for annotation are expensive, time-consuming, and prone to human error \cite{DBLP:journals/jcisd/RusinkoRFBKG24,liu2020drugcombdb}.
%That is, molecules can be viewed as graphs, and properties treated as preferences or labels, making the task analogous to recommending which molecules are likely to show a property, such as toxicity.
%, solubility, or therapeutic activity.
%However, just as recommender systems struggle with the cold-start problem for new users or items, molecular property prediction faces severe data scarcity.
%  and ROSIE\footnote{\url{https://rosie.rosettacommons.org/}}

Recently, few-shot learning has emerged as a promising paradigm for solving MPP in data-scarce scenarios \cite{DBLP:conf/sdm/MengLZ0K23,DBLP:conf/nips/0006LSW024,DBLP:conf/ijcai/ZhuangZWDFC23,DBLP:conf/cvpr/KimKKY19,DBLP:conf/www/GuoZYHW0C21}.
Most few-shot MPP approaches adopt a meta-learning framework with episodic training, where models are trained across a distribution of few-shot tasks with the goal of generalizing to unseen properties.
%tasks involving novel properties from only a handful of labeled molecules.
These methods can be categorized into two groups: traditional meta-learning approaches and in-context learning approaches.
Traditional approaches, e.g., optimization-based \cite{DBLP:conf/icml/FinnAL17,DBLP:conf/icml/AbbasXCCC22} and metric-based strategies \cite{DBLP:conf/nips/SnellSZ17}, focus on acquiring transferable knowledge for rapid adaptation to unseen tasks.
%That is, optimization-based methods, e.g., MAML, aim to learn parameter initializations that can be quickly fine-tuned to new tasks, while metric-based methods instead learn a discriminative latent space where molecular representations capture property-specific relationships.

% (1) Traditional methods include:
% MAML \cite{DBLP:conf/icml/FinnAL17},
% Sharp-MAML \cite{DBLP:conf/icml/AbbasXCCC22},
% ProtoNet \cite{DBLP:conf/nips/SnellSZ17},
% EGNN \cite{DBLP:conf/cvpr/KimKKY19}, and
% Meta-MGNN \cite{DBLP:conf/www/GuoZYHW0C21}.
% (2) In-context learning methods include:
% PAR \cite{DBLP:conf/nips/WangAYD21},
% GS-Meta \cite{DBLP:conf/ijcai/ZhuangZWDFC23},
% TPN \cite{DBLP:conf/ijcai/MaBALLZL20},
% HSL-RG \cite{DBLP:journals/nn/JuLQFWG0023}, and
% Pin-Tuning \cite{DBLP:conf/nips/0006LSW024}.

Since many molecular properties are interdependent, recent studies increasingly adopt in-context learning strategies to exploit relationships among molecules and seen properties for predicting unseen properties \cite{DBLP:conf/ijcai/ZhuangZWDFC23,DBLP:conf/nips/0006LSW024}.
That is, the model learns the relationship between a few seen molecular examples and properties and then leverages them as contextual signals to predict unseen properties.
For example, Pin-Tuning \cite{DBLP:conf/nips/0006LSW024} builds a context graph to represent molecule–property relationships and performs information propagation to derive property-aware molecular representations, improving the prediction of unseen properties.

However, two main challenges limit recent few-shot MPP methods.
\textbf{(1)} Recent studies mainly focus on relationships between molecules and properties, overlooking the chemical knowledge about functional groups that determine specific properties.
As shown in Figure~\ref{fig:problem}(a), only group $f_4$ serves as the causal substructure responsible for both molecules $m_2$ and $m_3$ exhibiting property $p_2$, highlighting the insufficiency of modeling at the molecule–property relationship alone.
We argue that capturing the interactions among functional groups, molecules, and properties is fundamental for achieving accuracy and interpretability in few-shot MPP tasks.
\textbf{(2)} It is challenging to identify a key substructure that is directly responsible for a particular property.
That is, a single molecule often comprises multiple functional groups, many of which may correlate with different properties.
However, for a specific property, only a subset of these substructures, the causal substructure, influences the property, while the remaining substructures are non-informative.
As shown in Figure~\ref{fig:problem}(b), consider the molecule 4-nitrophenol (\ce{C6H4(NO2)(OH)}), which contains both a hydroxyl group (\ce{OH}) and a nitro group (\ce{NO2}).
The \ce{OH} group contributes to solubility, whereas the \ce{NO2} group is the key causal determinant of toxicity.
Nevertheless, recent methods learn representations by aggregating all substructures, without disentangling key substructures from non-informative parts.
As a result, noisy information is incorporated into the graph-level representation, which can hinder the generalization in few-shot scenarios.

To discover important substructures, recent strategies have applied causal inference to graph learning tasks \cite{DBLP:conf/kdd/SuiWWL0C22,DBLP:conf/nips/0002ZB00XL0C22}.
%These strategies can be divided into two main groups: intervention-based methods that mitigate spurious correlations and invariance-based methods that discover stable subgraphs across environments.
%These methods can be grouped into two main categories: intervention-based approaches, which mitigate spurious correlations by disentangling causal from non-causal features, and invariance-based approaches, which aim to identify stable subgraphs across environments.
These methods can be grouped into two main categories: intervention-based approaches and invariance-based approaches.
The former seeks to mitigate spurious correlations by disentangling causal from non-causal features, while the latter aims to determine stable subgraphs across environments.
For example, CAL \cite{DBLP:conf/kdd/SuiWWL0C22} discovers causal patterns and reduces reliance on noisy information by estimating causal and non-causal components with attention modules, applying node-level masking strategies, and backdoor adjustment with noisy interveners.
In contrast, GIL \cite{DBLP:conf/nips/0002ZB00XL0C22} employs a GNN-based subgraph generator to extract invariant subgraphs while leveraging variant subgraphs to infer environment labels.
%, thereby enhancing generalization under distribution shifts.
%In contrast, invariance-based approaches aim to capture stable causal patterns by discovering invariant subgraphs across environments.
%For instance, GIL \cite{DBLP:conf/nips/0002ZB00XL0C22} employs a GNN-based subgraph generator to extract invariant subgraphs while leveraging variant subgraphs to infer latent environment labels, thereby enhancing generalization under distribution shifts.
However, two key challenges remain: 
\textbf{(1)} current methods often overlook relationships among functional groups, which are critical for identifying causal substructures that drive molecular properties and
\textbf{(2)} most studies rely on random interveners or augmentation-based schemes, which lack semantic grounding and thus fail to provide meaningful guidance for learning causal representations.

In this paper, we assume that each molecule contains a causal substructure that determines its target property, which is context-dependent and influenced by chemical knowledge about functional groups.
Unlike previous studies that treat molecules independently, we propose that causal substructures emerge from the relational context linking functional groups to properties, as shown in Figure~\ref{fig:problem}(a).
To achieve this, we construct a context graph that explicitly encodes these multi-level relationships to guide the discovery of causal substructures.
Then, we construct a Structural Causal Model (SCM) that identifies the causal substructure $C$ responsible for the target property $Y$.
%Last, we introduce a novel distribution intervention by combining $C$ with real-world chemical variations, disentangling true causal effects from confounding patterns.
By using the context graph as contextual signals and integrating graph causality inference, CaMol could estimate the true causal effect of causal substructure $C$ on the target property $Y$ in the few-shot MPP.

Building upon the SCM, we propose CaMol, a novel context-aware graph causality framework for few-shot MPP.
CaMol is designed to maximize the causal effect of the discovered causal substructure $C$ on the target property $Y$ by integrating contextual signals and graph causality inference.
CaMol is characterized by three main contributions.
\textbf{(1)} We construct a context graph that encodes the relationships among functional groups, molecules, and properties.
This context graph provides contextual guidance for discovering causal substructures, effectively addressing the few-shot challenge of limited labels and chemical priors for causality reasoning.
\textbf{(2)} We propose a learnable masking mechanism to separate causal substructures $C$ from spurious substructures $S$.
By selectively masking irrelevant atoms, CaMol focuses on key important substructures that are truly responsible for target properties.
\textbf{(3)} We propose a distribution intervention that parameterizes the backdoor adjustment by pairing causal substructures $C$ with chemically grounded confounders $S$ sampled from remaining molecules.
% a distribution intervention module that parameterizes the backdoor adjustment by pairing $C$ with chemically grounded confounders $S$, enabling the model to remove spurious dependencies and estimate faithful causal effects.
This enables the model to eliminate spurious dependencies and faithfully estimate the causal effect of $C$ on $Y$.
These components allow CaMol to discover interpretable and transferable causal substructures, achieving strong generalization in few-shot MPP tasks.

\section{Related Work}

%We now discuss how existing strategies address the few-shot problem in molecule prediction tasks compared to our proposed model.
%\subsection{Optimization-based Methods}
%IterRefLSTM and Meta-MGNN use graph-based molecular encoders to encode molecules regardless of target properties.
%\subsection{Metric-based Methods}

%Traditional methods adopt meta-learning techniques with a uniform episode sampling strategy during training. 
%These approaches can be mainly categorized into two groups: traditional methods and in-context methods. 

\subsection{Few-shot Molecular Property Prediction}

Traditional few-shot strategies primarily rely on optimization-based and metric-based learning to fine-tune parameter initialization in the latent space \cite{DBLP:conf/cvpr/KimKKY19,doi:10.1021/acscentsci.6b00367,DBLP:conf/nips/VinyalsBLKW16}.
For example, Meta-MGNN \cite{DBLP:conf/www/GuoZYHW0C21} improves MAML \cite{DBLP:conf/icml/FinnAL17} with self-attentive task weight meta-learning for few-shot learning.
Meta-MGNN incorporates auxiliary self-supervised objectives, such as bond reconstruction and atom type prediction, which are jointly optimized with the MPP tasks to strengthen representation learning.
%IterRefLSTM \cite{DBLP:journals/corr/Altae-TranRPP16} represents an early attempt to apply metric-based learning in few-shot MPP.
%It adapts Matching Networks \cite{DBLP:conf/nips/VinyalsBLKW16} to molecular property prediction, pioneering the use of metric learning in this domain.
The objective of EGNN \cite{DBLP:conf/cvpr/KimKKY19} is to learn the similarity of representations by exploiting intra-cluster similarity and inter-cluster dissimilarity to enhance few-shot performance.
However, the traditional methods overlook contextual and causal relations between molecules and properties, which limits their ability to generalize to unseen properties.

%Following this, Meta-GGNN [Meta-learning gnn initializations for low-resource molecular property prediction] and Meta-MGNN [Few-shot graph learning for molecular property prediction] introduce meta-learning with graph neural networks, setting a foundational framework that subsequent studies have continued to build upon.
%It is noteworthy that Meta-MGNN employs a pre-trained molecular encoder and achieves superior results through fine-tuning in the meta-learning process compared to training from scratch. 
%Meta-GNN employed a classic meta-learning method, MAML [ Model-agnostic meta-learning for fast adaptation of deep networks], with self-supervised tasks including bond reconstruction and atom type prediction.

%\subsection{Transductive Inference}

Recent strategies have been used in-context learning to guide the prediction of unseen properties by using contextual signals between molecules and properties \cite{DBLP:conf/sdm/MengLZ0K23,DBLP:conf/nips/0006LSW024,DBLP:conf/ijcai/ZhuangZWDFC23}.
These methods could be systematized into two main groups: 
molecule–molecule relational modeling and property-aware relational modeling.
The first strategies focus on leveraging molecular similarity to improve generalization.
PAR~\cite{DBLP:conf/nips/WangAYD21} builds a homogeneous context graph to connect similar molecules and refines embeddings through class prototypes and label propagation.
%HSL-RG~\cite{DBLP:journals/nn/JuLQFWG0023} applies graph kernels to construct relation graphs, enabling global communication of structural knowledge across neighboring molecules.
%MHNfs~\cite{DBLP:conf/iclr/SchimunekSFKRHK23} enriches few-shot tasks by incorporating a large-scale molecular library as contextual knowledge.
The second line emphasizes molecule–property dependencies to strengthen meta-learning.
GS-Meta~\cite{DBLP:conf/ijcai/ZhuangZWDFC23} constructs a molecule–property relation graph and redefines episodes as subgraphs of this graph, capturing richer property relations.
However, existing methods suffer from two key limitations: they overlook relations of functional groups as causal priors to properties, and they fail to discover key substructures that determine properties.

\subsection{Causal Inference in Graph Learning}

Recent studies have extended causal inference to graph learning, aiming to improve representation quality and robustness for downstream tasks~\cite{DBLP:conf/cvpr/YuLH23,DBLP:conf/nips/ZhuangZDBWLCC23,DBLP:conf/nips/ChenBZXHC23,DBLP:conf/nips/WuC0ZL24,DBLP:conf/www/FangLSGZWW024}.
These strategies fall into two research lines: (i) mitigating spurious correlations via interventions and (ii) discovering invariant causal subgraphs across environments.
The former methods primarily focus on mitigating spurious correlations by enforcing robustness against noisy features, while invariant-subgraph methods aim to extract stable causal patterns that generalize across environments.
The first line aims to reduce the effect of noisy substructures by applying interventions and enforcing invariance, thereby uncovering stable causal signals.
For example, CAL~\cite{DBLP:conf/kdd/SuiWWL0C22} disentangles causal from non-causal components using attention, distills causal parts through a node-masking strategy, and applies backdoor adjustment with non-causal interveners.
DIR~\cite{DBLP:conf/nips/Fan0MST22} similarly enforces robustness by identifying causal subgraphs that remain stable across environments and minimizing interventional risk variance.

In contrast, invariance-based approaches aim to identify subgraphs that remain stable across environments, capturing causal patterns while separating environment-specific variations~\cite{DBLP:conf/nips/ZhuangZDBWLCC23,DBLP:conf/nips/0002ZB00XL0C22,DBLP:conf/icml/LiWZW0C22}.
For example, GIL~\cite{DBLP:conf/nips/0002ZB00XL0C22} uses a GNN-based subgraph generator to extract invariant subgraphs while using variant ones to infer latent environment labels.
The idea of RGCL~\cite{DBLP:conf/icml/LiWZW0C22} is to perform probabilistic sampling over important nodes to highlight causal substructures, while unimportant nodes guide augmentation and align contrastive learning with causal rationale discovery.
%DIR \cite{DBLP:conf/nips/Fan0MST22} identifies causal subgraphs by employing a node-masking matrix parameterized by an MLP and sigmoid function, while treating non-causal features as noisy interveners to construct intervention pairs.
However, most invariance-based methods ignore contextual signals from functional groups, which are essential for capturing the true causal patterns that determine molecular properties.
Moreover, the random or augmented interveners could introduce unrealistic confounding substructures, limiting the reliability of inference across environments.

\section{Problem Description}

\subsection{Few-shot Molecular Property Prediction}

Let $\mathcal{T}$ be a set of tasks, where each task $T \in \mathcal{T}$ corresponds to predicting a molecular property $y$.  
The training set is given by $\mathcal{D}_{\text{train}} = \{(m_i, y_{i,t}) \mid t \in \mathcal{T}_{\text{train}}\}$, while the test set $\mathcal{D}_{\text{test}}$ contains tasks with disjoint property sets, i.e., $\mathcal{P}{\text{train}} \cap \mathcal{P}{\text{test}} = \varnothing$.

The goal of few-shot MPP is to learn from $\mathcal{D}_{\text{train}}$ and generalize to unseen tasks in $\mathcal{D}_{\text{test}}$ with only a few labeled molecules. 
Following standard meta-learning, we adopt episodic training \cite{DBLP:conf/icml/FinnAL17}.
Typically, the task is formulated as a binary classification, where molecules are labeled as active ($y=1$) or inactive ($y=0$).
Thus, a 2-way $K$-shot episode $E_t = (\mathcal{S}_t, \mathcal{Q}_t)$ is constructed, where $\mathcal{S}_t$ refers to the support set,
and $\mathcal{Q}_t$ is the query set used for evaluation.

%each episode $E_t = (\mathcal{S}_t, \mathcal{Q}_t)$ samples a task $T_t$ with support set $\mathcal{S}_t = \{(m^s_i, y^s_{i,t})\}_{i=1}^{2K}$ containing $K$ molecules per class, and query set $\mathcal{Q}_t = \{(m^q_i, y^q_{i,t})\}_{i=1}^{M}$ used for evaluation.

\subsection{A Causal View in Molecular Graphs}

We formalize the causal relationships among the substructures of a molecular graph by using a Structural Causal Model (SCM), as shown in Figure \ref{fig:Few_shot}.
Specifically, we consider the input molecular graph $G$, which consists of two disjoint substructures: the causal substructure $C$, which is directly responsible for the molecular property, and the noisy substructure $S$.  
We define the SCM as follows, where each link denotes a relationship:

\begin{figure}[tb]
\centering
  \includegraphics[width=  \linewidth]{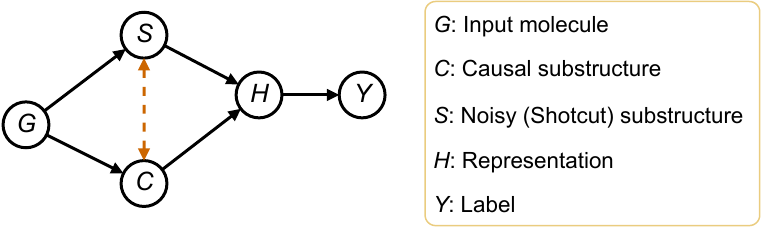}
  \caption{Causal relationships between variables in MPP.}
  \label{fig:Few_shot}
\end{figure}

\begin{itemize}
    
    \item $C \leftarrow G \rightarrow S$. The input molecular graph $G$ is composed of two disjoint substructures, the causal substructure $C$ and the noisy substructure $S$.
    
    \item $C \rightarrow H \rightarrow Y$. The causal substructure $C$ generates representations $H$ that determine the ground-truth property $Y$.  
    
    % \item $S \rightarrow H \rightarrow Y$. The noisy substructure $S$ correlates spuriously with $Y$, hindering the true causal effect of $C$.  
    
    \item $C \dashleftrightarrow S$. This denotes a spurious correlation between $C$ and $S$, which may arise from direct dependence or from unobserved confounders.
    %\item $C \longleftrightarrow S$. The correlation between the causal substructures $C$ and non-informative substructures $S$.

\end{itemize}

Based on our SCM, we identify a backdoor path between $C$ and $Y$, i.e., $S \leftarrow G \rightarrow C \rightarrow H \rightarrow Y$, where $S$ represents the confounding substructure that introduces spurious correlations between $C$ and $Y$.
To accurately estimate the causal effect of $C$ on $Y$, it is necessary to block this backdoor path by removing the influence of $S$, which is the only variable satisfying the backdoor criterion.
This ensures that the prediction of $Y$ depends solely on the causal substructure $C$, rather than on the confounding substructure $S$.
The environmental variable $S$ thus acts as a confounder between $G$ and $Y$, potentially introducing noisy correlations between $C$ and $Y$.

%To ensure that the prediction $Y$ is uncorrelated with $S$, it is necessary to eliminate the confounding effect of $S$, which is the only remaining element that satisfies the backdoor criterion.  
%This encourages the model to rely on the causal substructure $C$ together with the molecular graph $G$ for predicting the label $Y$.  
%if no direct causal link $C \rightarrow Y$ exists.  
%Thus, severing the dependence $C \dashleftrightarrow S$ is crucial to ensure that predictions are based solely on the true causal effect of $C$.  

%$C \longleftrightarrow S$ can create spurious correlations between the non-causal substructures $S$ and the ground-truth label $Y$.

\begin{figure*}[t]
\centering
  \includegraphics[width=  \linewidth]{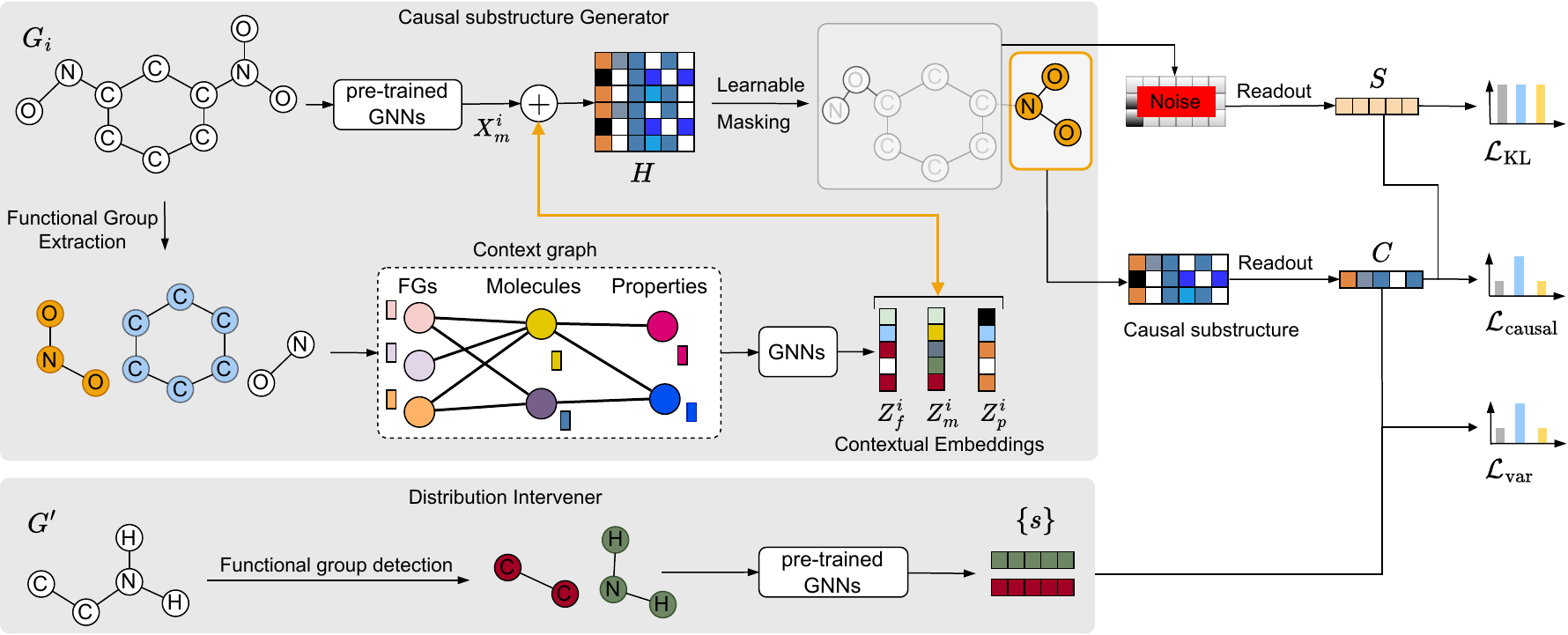}
\caption{Overview of the CaMol architecture, consisting of a causal substructure extractor and a distribution intervener.}
  \label{fig:architecture}
\end{figure*}

\section{Methodology}

% \subsection{A Causal View in Molecular Graphs}
% \subsection{Context Graph Reconstruction}

% \subsection{Causal Learner for Molecular Graphs}
% \subsubsection{Causal Substructure Generator via Atom Masking}
% \subsubsection{Backdoor Adjustment}
% \subsubsection{Distribution Intervention}

% \subsection{Meta Training}

% \subsection{Theoretical Analysis}

%\section{Causal Molecular Learner}

%\section{Causal Substructure Generator}

\subsection{Context Graph Learning}

To capture the relational structure between molecules and their properties, we construct a context graph that integrates support and query molecule samples within each episode.
Specifically, we decompose each molecule into functional groups using the BRICS algorithm \cite{degen2008art}, which uses domain knowledge to fragment molecules into chemically meaningful subgraphs.
%Specifically, given an input molecule $m$, we define the set of functional group fragments as $F = \{ G^{(0)},G^{(1)}, \cdots, G^{(m)} \}$, where $G^{(i)} = (V^{(i)}, E^{(i)})$ denotes the $i$-th subgraph and $k$ is the number of fragments.
Given an episode $E_t$, the context graph is denoted as $G_t = (V_t, A_t, X_t)$.
The node set $V_t$ contains three types of nodes:
$\{M\}$ molecule nodes,
$\{F\}$ functional group nodes, and
$\{P\}$ property nodes.
Three types of edges are used to represent relationships: positive-label edges, negative-label edges, and unknown-label edges, indicating whether the given molecule is active, inactive, or unknown w.r.t. the property, respectively.
This context graph explicitly encodes domain knowledge and relational signals, providing contextual guidance for discovering causal substructures in few-shot MPP.
%This graph-based representation models the co-occurrence and interaction of substructures and their associated properties within a task, enabling the learner to leverage relational information for more effective few-shot inference.
We then learn the node representations with a GNN-based encoder as:
\begin{eqnarray}
Z = \text{GNN}(V_t, A_t, X_t), 
\end{eqnarray}
where $Z \in \mathbb{R}^{(M + F + P)\times d}$ denotes the learned context representation matrix for $E_t$ and
$\text{GNN}(\cdot)$ is a GNN encoder, e.g., EGIN \cite{DBLP:conf/iclr/HuLGZLPL20}.
Here, $V_t$ and $A_t$ denote the node set and adjacency matrix of the context graph, respectively, and $X_t$ denotes the initial features of the nodes.  
The features of molecules are initialized by the pre-trained molecular encoder S-CGIB~\cite{Hoang_Lee_2025}, and property nodes are randomly initialized, following the work \cite{DBLP:conf/nips/0006LSW024}.

We note that the context embedding $Z$ contains functional group embeddings $Z_f$, molecule embeddings $Z_m$, and property embeddings $Z_p$.
Then, for each input molecule $m_i$, we obtain its contextual representations by concatenating its representations with the corresponding context embeddings, as:
\begin{eqnarray}
H_i = \left( X^{i}_m \; \Big\| \; Z^{i}_f \; \Big\| \; Z^{i}_m \; \Big\| \; Z^{i}_p \right) ,
\end{eqnarray}
where $X^{i}_m$ denotes the node-level representation of molecule $m_i$ obtained from the pre-trained encoder S-CGIB, 
%$Z^{i}_f$, $Z^{i}_m$, and $Z^{i}_p$ refer to the embeddings of the functional groups, molecule, and property from the context graph, respectively, 
and $\|$ denotes vector concatenation. 
The output node embedding $H$ is then used for causal inference, as represented in the following sections.
%When predicting the target property $p$ of molecule $m$, we take the learned representations of this molecule $c_m$ and of the target property $c_p$ as the context vectors.

\subsection{Graph Causality Learner for Molecules}

\subsubsection{Backdoor Adjustment}

% A key challenge in molecular property prediction is the presence of confounding substructures, as S, which may correlate with the target property Y without being causally responsible. 
% For example, aromatic rings may frequently co-occur with toxic functional groups but do not themselves cause toxicity. These shortcut patterns can bias the model if not properly accounted for.

% One straightforward way to alleviate the confounding effect of S is to
% collect molecules with a causal substructure C and various shortcut substructures $\backslash S$, which is impossible due to its computational complexity.
% To address this issue without requiring exhaustive data collection, we adopt a causal inference perspective. 
% By treating the presence of C as an intervention, we aim to isolate its effect on 
% Y by assuming or estimating independence between C and S in the interventional distribution, as:
%For example, aromatic rings may frequently co-occur with toxic functional groups but do not themselves cause toxicity. 
%However, this is computationally infeasible due to the combinatorial nature of chemical space.
%The symbol $\tilde{P}$ is used to denote the post-intervention distribution, distinguishing it from the observational distribution $P$. 

A main challenge in discovering important substructures is the presence of confounding substructures $S$ that correlate spuriously with the label $Y$, hindering the true causal effect of $C$.
A direct solution would be to collect molecules including $C$ across all possible variations of $S$ from the dataset, but this is infeasible due to the difficulty of acquiring sufficiently diverse and annotated data, especially in few-shot scenarios.
% A straightforward solution is to collect all molecules containing a causal substructure $C$ across all variations of confounding effects from $S$ and learn the correct patterns.
% However, this is impractical due to the difficulty of acquiring sufficiently diverse and annotated molecular data in few-shot scenarios.
% \begin{eqnarray}
% P\left( Y \mid \textbf{do}(C) \right) 
% &=& \tilde{P}(Y \mid C) \nonumber \\
% &=& \sum_s \tilde{P}(Y \mid C, s) \cdot \tilde{P}(s \mid C) 
% \quad  \nonumber \text{(Bayes’ Rule)} \\  
% &=& \sum_s \tilde{P}(Y \mid C, s) \cdot \tilde{P}(s) 
%  \nonumber \   \quad \text{(Assume } S \perp C ) \ ,
% \end{eqnarray}
Causal theory offers a principled solution via the backdoor adjustment \cite{DBLP:conf/iclr/WuWZ0C22}.
Instead of modeling the confounded distribution directly, the model estimates the intervention distribution, as:
\begin{eqnarray}
P\left( Y \mid \textbf{do}(C) \right) 
&=& \tilde{P}(Y \mid C) \nonumber \\
&=& \sum_s \tilde{P}(Y \mid C, s) \cdot \tilde{P}(s \mid C) 
\quad  \nonumber \\  
&=& \sum_s \tilde{P}(Y \mid C, s) \cdot \tilde{P}(s) 
 \nonumber \   ,
\end{eqnarray}
where $\textbf{do}(C)$ represents an intervention that sets the causal substructure $C$ to a fixed value.
Under the assumption that $S$ is independent of $C$, i.e., $S \perp C$, the intervention distribution $P(Y \mid \textbf{do}(C))$ can be computed by marginalizing over $S$ using the law of total probability.
By doing so, the backdoor adjustment enables the model to make robust and invariant predictions by learning from various confounding substructures.
%Unlike prior studies, we further (1) refine the intervention space using real-world semantic substructure confounders and (2) use a context graph that incorporates prior chemical knowledge to guide the causal inference.
Unlike previous studies, CaMol extends this idea by (1) refining the intervention space $s$ using semantic substructure confounders that are grounded in chemically meaningful functional groups and
(2) incorporating a context graph that encodes prior chemical knowledge to guide the causal inference.
%The refined intervention space prevents the inclusion of unrealistic confounders, while the context graph provides relational guidance that anchors causal inference in chemically interpretable patterns.
%We argue that the context graph and semantic intervention are one of the key successes of CaMol.
The theoretical analysis is provided in Appendix \ref{app:theory_1}.

\subsubsection{Causal Substructure Extractor via Atom Masking}
%Instead of manipulating the molecular graph $G$ directly—which could violate chemical validity—
%Specifically, we operate on the atom-level embedding matrix $H$, which encodes structural and chemical information from $G$ as learned by the graph encoder (e.g., GIN).
%to identify and isolate the causal atoms, we introduce a learnable importance estimation mechanism. 
%, which may correlate with the label but do not truly determine the property.
%This importance score $p_i$ reflects how much atom $v_i$ contributes to the model prediction $Y$, as:

To incorporate a causal intervention framework into few-shot MPP, a key challenge is separating the causal substructure $C$, which directly determines the target property, from confounding substructures $S$.
In molecular graphs, however, this separation is challenging as directly removing substructures could violate chemical validity.
To address this, we introduce a learnable masking strategy that disentangles atom-level causal substructures from non-informative ones.
Specifically, for each atom $v_i$, we estimate its relevance to the prediction task with a multilayer perceptron (MLP), as:
\begin{eqnarray}
& {{p}_{i}}=\text{MLP}\left( H_{i} \right) ,\
\end{eqnarray}
where $H_i$ is the atom representation.
Based on $p_i$, we derive atom-level causal and confounding representations through a noise-injected masking scheme, as:
\begin{eqnarray}
& C_{i} &= {{\lambda }_{i}}H_{i}+\left( 1-{{\lambda }_{i}} \right)\varepsilon  , \\ 
& S_{i} &= \left( 1-{{\lambda }_{i}} \right)H_{i} , \
\end{eqnarray}
where $\lambda_i \sim \text{Bernoulli}(p_i)$,
$\varepsilon \sim \mathcal{N}(\mu_H, \sigma_H^2)$ is sampled noise, and $\mu_H$ and $\sigma_{H}$ are the mean and variance of $H$, respectively.
By doing so, the model could disentangle causal substructures by selectively masking unimportant atoms with noisy information with $p_i$.

% where ${{\lambda }_{i}} \sim \operatorname{Bernoulli}\left( {p_i} \right)$ is obtained by sampling from a Bernoulli distribution parameterized with the probability $p_i$, 
% ${{\mu }_{{H}}}$ and ${{\sigma }_{{H}}}$ are the mean and variance of ${H}$,
% $\varepsilon \sim N\left( {{\mu }_{{H}}} \right)$ is sampled from ${H}$ based on Gaussian distribution.
%Thus, the important substructure is disentangled by masking unimportant nodes with noisy information, with the probability of $p_i$.
%input graph is compressed into $Z$ with the probability of $p_i$ by masking unimportant nodes with noisy information.
%That is, the compression process ensures that $Z$ focuses on the important nodes (graph core) while discarding irrelevant and unimportant nodes.
% To enable differentiable sampling of substructures, we adopt the Gumbel–Sigmoid reparameterization trick~\cite{DBLP:conf/iclr/JangGP17,DBLP:conf/iclr/MaddisonMT17}, which allows us to approximate discrete Bernoulli sampling in a continuous and differentiable manner. Specifically, the sampling of $\lambda_i$ for each atom $v_i$ is computed as:
% \begin{eqnarray}
% \lambda_i = \operatorname{Sigmoid} \left(\frac{1}{\tau} \bigg[ \log\frac{p_i}{1-p_i} + \log \frac{q}{1-q} \bigg]\right),
% \end{eqnarray}
% where $q \sim \operatorname{Uniform}(0,1)$ and $\tau$ is the temperature parameter that controls the smoothness of the approximation.
To enable differentiable sampling, we adopt the Gumbel–Sigmoid strategy, i.e.,  ${{\lambda }_{i}}=\operatorname{Sigmoid}( 1/\tau \log [ {{p}_{i}}/(1-{{p}_{i}}) ] )+\log \left[ q/(1-q) \right]$ where $ q \sim \operatorname{Uniform}(0,1)$ and $\tau$ is the temperature parameter \cite{DBLP:conf/iclr/JangGP17,DBLP:conf/iclr/MaddisonMT17}.
%With the learned weights, we disentangle the causal representation $C$ and the confounding representation $S$.
Finally, graph-level embeddings for $C$ and $S$ are obtained via a summation readout function over their node representations.
%This formulation ensures that the masking strategy remains fully differentiable, enabling end-to-end training of CaMol with gradient-based optimization.
% \begin{eqnarray}
% & p_i = \operatorname{Sigmoid} \left ( \operatorname{MLP}({{H}_i}) \right ) ,\\ 
% & \varepsilon \sim N\left( {{\mu }_{{H}}},{{\sigma }^2_{H}} \right), \ 
% {{\lambda }_{i}} \sim \operatorname{Bernoulli}\left( {p_i} \right), \label{eq:lambda}\\
% & {{z}_{i}}={{\lambda }_{i}} {h_i}+(1-{{\lambda }_{i}})\varepsilon  ,
% \end{eqnarray}
%To model the causal effect of $C$, we minimize the aggregated $s$-interventional risks over the dataset as:

\subsubsection{Distribution Intervention}

%A direct approach to backdoor adjustment would be to explicitly generate intervened molecular structures, i.e., molecules that contain the identical substructure $C$ but vary in noisy substructures $S$ across the dataset.
%However, directly generating such structures is non-trivial, as molecules must obey chemical validity and are governed by complex domain-specific rules.

We aim to disentangle causal substructures $C$ whose relationship with the label $Y$ remains invariant despite the presence of noisy confounders $S$ across molecules.
To address this, we employ the backdoor adjustment to mitigate the influence of $S$, which operates implicitly in the representation space.
By partitioning the confounders $S$ into distinct groups $S = \{s\}$, we construct multiple $s$-interventional distributions that help isolate the causal effect of $C$ on the target property $Y$.
The invariance loss can be defined as:
\begin{eqnarray}
\mathcal{L}_{\text{var}} &=& \sum\limits_{G \in \mathcal{D}_{\text{train}}}{\sum\limits_{s}{ \mathcal{L} \left( Y,C, s \right)}}  \nonumber \\ 
&=&\sum\limits_{G \in \mathcal{D}_{\text{train}}} \sum\limits_{s \in S} \left[ - y \log \hat{y}_{C, s} - (1 - y) \log (1 - \hat{y}_{C, s}) \right], 
\label{eq:loss_var}
\end{eqnarray}
where $\hat{y}_{C, s} = f_\theta(C, s)$ denotes the prediction derived from the causal representation $C$ conditioned on the confounder $s$.
Unlike prior studies using augmentation or noise injection, CaMol leverages functional group knowledge to guide interventions.
Confounding candidates $\{s\}$ are extracted from functional groups in graphs $G'$ linked to properties different from the target property.
%By doing so, CaMol ensures that the interventions are semantically valid while improving robustness by preventing the model from relying on spurious correlations.

\subsubsection{Objective Functions}

A principle of our framework is that the confounding substructure $S$ carries no predictive information about the label $Y$.
Thus, we regularize the model such that predictions based on $S$ follow a non-informative prior, i.e., a uniform distribution over the label space.
%That is, we use the KL divergence to push the model's predictions from $S$ toward being uninformative, i.e., uniform. 
Formally, we minimize the KL divergence between the predictive distribution from $S$ and a uniform distribution from the label space, as:
\begin{eqnarray}
\mathcal{L}_{\text{KL}} &=& \sum\limits_{G \in \mathcal{D}_{\text{train}}}{ { \mathcal{L} \left( Y_{\text{random}}, S \right)}}  \nonumber \\
&=& \sum\limits_{G \in \mathcal{D}_{\text{train}}} \text{KL} \left( P_\theta(Y \mid S) \; \Big\| \; \mathcal{U}(Y) \right),
\label{eq:loss_kl}
\end{eqnarray}
where $P_\theta(Y \mid S)$ is the predicted label distribution from confounding representations and $\mathcal{U}(Y)$ is a uniform distribution.

In contrast, we enforce the causal substructure $C$ to retain predictive information about the label.
To achieve this, we optimize a causal prediction loss using cross-entropy:
\begin{eqnarray}
\mathcal{L}_{\text{causal}} &=& \sum\limits_{G \in \mathcal{D}_{\text{train}}}{ {\mathcal{L}\left( Y, C \right)}}   \nonumber \\
 &=& \sum\limits_{G \in \mathcal{D}_{\text{train}}} \left[ - y \log \hat{y}_C - (1 - y) \log (1 - \hat{y}_C) \right],
\label{eq:loss_causal}
\end{eqnarray}
where $\hat{y}_C = f_\theta(C,S)$ is the prediction based on the causal substructure $C$ and the noise $S$. 
The final loss function is defined as:
%Finally, the model is trained end-to-end with a joint objective that balances causal prediction, confounder suppression, and intervention regularization:
\begin{eqnarray}
& \mathcal{L}_{\text{tot}} = \mathcal{L}_{\text{causal}}  + \alpha_1 \mathcal{L}_{\text{KL}} + \alpha_2 \mathcal{L}_{\text{var}} \ , 
\end{eqnarray}
where $\alpha_1$ and $\alpha_2$ are hyperparameters.

\subsection{Meta-training Optimization}

Following the MAML framework \cite{DBLP:conf/icml/FinnAL17}, we adopt a gradient-based meta-learning strategy.
A batch of $B$ episodes $\{ \mathcal{E}_t \}_{t=1}^{B}$ is randomly sampled.
For each episode $\mathcal{E}_t$, the inner-loop optimization is performed on the support set $\mathcal{S}_t$ by computing the total loss:
\begin{eqnarray}
%\mathcal{L}_{\text{inner}} (f_{\theta}) = - \sum\nolimits_{  \mathcal{S}_t} \left[ y \log \hat{y} + (1 - y) \log(1 - \hat{y}) \right],
\mathcal{L}_{\text{S}} (f_{\theta})  = \frac{1}{B} \sum_{t=1}^{B} \left( 
\mathcal{L}_{\text{causal}}^{(t,S)} + 
\alpha_1 \mathcal{L}_{\text{KL}}^{(t,S)} + 
\alpha_2 \mathcal{L}_{\text{var}}^{(t,S)} 
\right)  , \ 
\label{eq:inner-loss}
\end{eqnarray}
where $f_\theta$ denotes the model parameterized by $\theta$.
%, and $\hat{y}$ is the predicted label. 
The model parameter $\theta$ is then updated via gradient descent:
\begin{eqnarray}
\theta' \leftarrow \theta - \alpha_{\text{S}} \nabla_\theta \mathcal{L}_{\text{S}} (f_{\theta}),
\label{eq:inner-update}
\end{eqnarray}
where $\alpha_{\text{inner}}$ is the inner-loop learning rate.

In the outer loop, the model update parameter $ {\theta'}$ is evaluated on the corresponding query set $\mathcal{Q}_t$, and the classification loss is computed as $\mathcal{L}_{\text{Q}}$. 
%The total meta-training objective across all $B$ tasks includes both the query loss and a regularization term $\mathcal{L}_{\text{Emb-BWC}}$:
The total meta-training objective across all $B$ tasks incorporates both the query-set causal loss, as:
\begin{eqnarray}
\mathcal{L}_{\text{Q}} (f_{\theta'})  = \frac{1}{B} \sum_{t=1}^{B} \left( 
\mathcal{L}_{\text{causal}}^{(t,Q)} + 
\alpha_1 \mathcal{L}_{\text{KL}}^{(t,Q)}  + 
\alpha_2 \mathcal{L}_{\text{var}}^{(t,Q)}
\right)  , 
\label{eq:outer-loss}
\end{eqnarray}
where $\lambda_1$ and $\lambda_2$ are hyper-parameters. 
% \begin{eqnarray}
% \mathcal{L}(f_{\theta'}) = \frac{1}{B} \sum_{t=1}^{B} \mathcal{L}^{\text{cls}}_{t, \mathcal{Q}}(f_{\theta'}) + \lambda \mathcal{L}_{\text{Emb-BWC}},
% \label{eq:meta-loss}
% \end{eqnarray}
The model parameters are updated using an outer-loop gradient descent step:
% \begin{eqnarray}
% \theta \leftarrow \theta - \alpha_{\text{outer}} \nabla_\theta \mathcal{L}(f_{\theta'}),
% \label{eq:outer-update}
% \end{eqnarray}
% where $\alpha_{\text{outer}}$ is the meta-learning rate.
\begin{eqnarray}
\theta \leftarrow \theta - \alpha_{\text{outer}} \nabla_\theta \mathcal{L}_{\text{Q}} (f_{\theta'}) , 
\label{eq:13}
\end{eqnarray}
where $\alpha_{\text{outer}}$ is the meta-learning rate. 
The training process is provided in Appendix B. % \ref{app:alg}.

\subsection{Computational Complexity Analysis}

CaMol introduces only a lightweight computational overhead compared to standard few-shot GNN baselines, e.g., MAML \cite{DBLP:conf/icml/FinnAL17}.
The molecular backbone uses a pre-trained S-CGIB encoder \cite{Hoang_Lee_2025}, whose cost is fixed and thus does not increase the training complexity of CaMol. 
In each episode, the context graph encoder processes $N_c$ nodes and $E_c$ edges with a complexity of $O\!\left(L_c (N_c + E_c)d\right)$, where $L_c$ is the number of GNN layers and $d$ is the embedding dimension.  
The causal substructure generator consists of two lightweight MLPs applied at the atom level, resulting in a complexity of $O\!\left( N d^2\right)$, where $N$ is the number of atoms in the molecule.
The distribution intervention module combines the causal representation with a chemically grounded confounder in the representation space, introducing only constant-time overhead per molecule.
Overall, CaMol follows a MAML-style meta-optimization scheme and has a per-iteration complexity of $O\!\left(B \big(L_c (N_c + E_c)d + A d^2\big)\right)$, where $B$ is the number of tasks per batch.

%We also provide the hy in Appendix \ref{tab:appendix_hyperparameters}.

\section{Experiments}

We evaluate CaMol on three tasks: few-shot performance, 
molecule interpretation, 
and model explainability.
%This verifies whether the causal substructures truly interpret the molecule by capturing the essential components responsible for its properties.
The first task analyzes whether the found causal substructures can enhance the predictive performance of few-shot MPP.
The second task evaluates whether the found causal substructures are consistent with preserving the similar property.
The third task is to evaluate whether the explanations are faithful to the model’s decision.

\begin{table*}[tb]
\centering
\caption{A performance comparison on few-shot MPP tasks in terms of ROC-AUC scores (\%).}
\setlength{\tabcolsep}{1.3 mm} % Default value: 6pt
\fontsize{9 pt}{9 pt}\selectfont
%\small
\begin{adjustbox}{width= \linewidth}
\begin{tabular}{l ccc ccc ccc ccc ccc}
\toprule
\multirow{2}{*}{Methods}
% Tox21 SIDER MUV ToxCast
&\multicolumn{3}{c}{Tox21}
&\multicolumn{3}{c}{SIDER} 
&\multicolumn{3}{c}{MUV} 
% &\multicolumn{3}{c}{ToxCast} 
% &\multicolumn{3}{c}{PCBA} 
\\
\cmidrule(lr){2-4}
\cmidrule(lr){5-7}
\cmidrule(lr){8-10}
% \cmidrule(lr){11-13}
% \cmidrule(lr){14-16}

&10-shot
&5-shot
&1-shot

&10-shot
&5-shot
&1-shot

&10-shot
&5-shot
&1-shot

% &10-shot
% &5-shot
% &1-shot

% &10-shot
% &5-shot
% &1-shot
\\ 
\midrule
MAML
&79.59$\pm$0.33 
&77.12$\pm$0.61
&75.63$\pm$0.18 

&70.49$\pm$0.54 
&70.05$\pm$0.87
&68.63$\pm$1.51

&68.38$\pm$1.27
&66.47$\pm$0.32
&65.82$\pm$2.49 

% &68.40$\pm$1.02
% &67.55$\pm$0.65
% &61.11$\pm$1.36

% &66.22$\pm$1.31
% &65.25$\pm$0.75
% &62.04$\pm$1.73

\\
Sharp-MAML
&75.37$\pm$0.23  
&75.09$\pm$0.72
&74.59$\pm$0.56 

&71.02$\pm$ 0.81  
&70.49$\pm$ 0.29
&68.43$\pm$ 0.96 

&65.52$\pm$2.01 
&65.37$\pm$3.33
&65.12$\pm$2.98 

% &66.95$\pm$1.42
% &65.41$\pm$0.95
% &63.76$\pm$1.06

% &73.19$\pm$2.37 
% &71.25$\pm$1.84
% &69.23$\pm$1.36
\\

ProtoNet
&72.99$\pm$0.56 
&72.04$\pm$2.13
&68.22$\pm$0.46

&61.34$\pm$1.08 
&60.88$\pm$1.84
&57.41$\pm$0.76 

&68.92$\pm$1.64
&64.86$\pm$2.31
&64.81$\pm$1.95

% &68.87$\pm$0.82
% &66.25$\pm$1.04
% &58.55$\pm$1.18

% &64.93$\pm$1.94
% &62.29$\pm$2.12
% &55.79$\pm$1.45

\\
EGNN
&80.11$\pm$0.31
&76.80$\pm$2.62
&75.71$\pm$0.21

&71.24$\pm$0.37
&69.14$\pm$1.22
&66.36$\pm$0.29

&68.84$\pm$1.35
&66.92$\pm$1.27
&62.72$\pm$1.97 

% &66.42$\pm$0.77
% &65.01 $\pm$0.94
% &63.98 $\pm$1.20

% &69.92 $\pm$1.85 
% &68.35 $\pm$1.52
% &67.71$\pm$ 3.67

\\

Meta-MGNN

&83.44$\pm$0.14
&83.03$\pm$0.25
&\underline{82.67 $\pm$0.20}

&77.84 $\pm$0.34
&76.19$\pm$0.65
&74.62$\pm$0.41

&68.31 $\pm$3.06
&67.91$\pm$ 2.09
&66.10$\pm$ 3.98

% &76.27$\pm$0.57
% &75.26$\pm$0.93
% &72.43$\pm$0.85

% &72.58$\pm$0.34
% &72.55$\pm$0.19
% &72.51$\pm$0.52
\\
 
\midrule

PAR
&82.13$\pm$0.26 
&81.92$\pm$0.37
&80.02$\pm$0.30 

&75.15$\pm$0.35 
&74.01$\pm$0.12
&72.33$\pm$0.47

&68.08$\pm$2.23 
&67.37$\pm$1.32
&65.62$\pm$3.49

% &74.77$\pm$0.85
% &71.25$\pm$1.17
% &69.45$\pm$1.34

% &78.35$\pm$2.03
% &76.15$\pm$1.19
% &74.02$\pm$1.68

\\
TPN
&76.05$\pm$0.24
&75.45$\pm$0.95
&60.16$\pm$1.18

&67.84$\pm$0.95
&66.52$\pm$1.28
&62.90$\pm$1.38

&65.22$\pm$5.82
&65.13$\pm$0.23
&50.00$\pm$0.51

% &69.47$\pm$0.71
% &66.04$\pm$1.14
% &63.78$\pm$0.05

% &67.61$\pm$0.33
% &63.66$\pm$1.64
% &60.28$\pm$1.39

\\ 

GS-Meta
&\underline{86.38$\pm$0.69} 
&84.13$\pm$0.72
&79.32$\pm$0.89 

&\underline{83.72$\pm$0.54}
&\underline{83.22$\pm$0.51}
&\underline{82.84$\pm$0.67 }

&67.11$\pm$1.95 
&64.50$\pm$0.20
&64.70$\pm$2.88

% &-
% &-
% &-

% &-
% &-
% &-
\\
HSL-RG
&77.51$\pm$0.62
&75.89$\pm$1.28
&72.04$\pm$0.79

&75.84$\pm$1.13
&73.07$\pm$0.87
&70.15$\pm$0.83

&69.44$\pm$1.09
&68.36$\pm$0.46
&\underline{66.62$\pm$0.59}

\\
Pin-tuning
&84.94$\pm$0.35
&\underline{84.89$\pm$1.08}
&80.76$\pm$0.43

&89.58$\pm$0.58
&83.15$\pm$0.24
&81.63$\pm$0.81

&\underline{72.49$\pm$0.21}
&\underline{71.78$\pm$0.24}
&63.54$\pm$0.77

% &84.75$\pm$1.41
% &83.03$\pm$1.05
% &76.98$\pm$1.92

% &80.52$\pm$0.59
% &78.35$\pm$0.64
% &75.98$\pm$0.83
\\

\midrule

% CaMol w/o M.
% &86.26$\pm$1.83  
% &84.95$\pm$0.91
% &83.15$\pm$2.94  

% &87.03$\pm$1.27  
% &86.15$\pm$0.75
% &81.07$\pm$3.01  

% &83.52$\pm$1.22  
% &80.22$\pm$0.82
% &73.41$\pm$0.73  
% \\

% CaMol w/o C.G.
% &79.37$\pm$1.12  
% &78.66$\pm$1.11
% &75.66$\pm$0.95

% &78.45$\pm$2.39  
% &77.7$\pm$0.82  
% &74.12$\pm$0.28

% &76.94$\pm$3.01  
% &75.18$\pm$0.92
% &72.57$\pm$2.17
% \\

CaMol (Ours)
&\textbf{92.69$\pm$0.49}
&\textbf{86.89$\pm$0.62}
&\textbf{83.67$\pm$0.38}

&\textbf{91.89$\pm$0.66} % &90.37$\pm$0.41
&\textbf{88.43$\pm$0.74} %&84.08$\pm$0.16
&\textbf{83.36$\pm$0.32} %&82.42

&\textbf{84.76$\pm$0.63} % &95.35$\pm$1.06
&\textbf{81.37$\pm$0.74} % &95.28$\pm$0.18
&\textbf{76.79$\pm$0.42} % &85.42$\pm$0.33 

% &89.64$\pm$0.25
% &87.46$\pm$0.29
% &82.65$\pm$0.68

% &87.74$\pm$0.59
% &84.92$\pm$0.24
% &78.26$\pm$0.41
\\

\bottomrule
% \end{tabular}
% %\end{adjustbox}
% \label{tab:kshot_1}
% \end{table*}

% \begin{table*}[t]
% \centering
% %\renewcommand*{\arraystretch}{0.80}
% \caption{Few-shot performance in terms of ROC-AUC scores (\%).}
% \setlength{\tabcolsep}{1 mm} % Default value: 6pt
% \fontsize{9 pt}{9 pt}\selectfont
% %\small
% %\begin{adjustbox}{width= \linewidth}
% \begin{tabular}{l ccc ccc ccc }
\toprule
\multirow{2}{*}{Methods}
% Tox21 SIDER MUV ToxCast
%&\multicolumn{3}{c}{Tox21} 
%&\multicolumn{3}{c}{SIDER} 
%&\multicolumn{3}{c}{MUV} 
&\multicolumn{3}{c}{ToxCast} 
&\multicolumn{3}{c}{PCBA} 
&\multicolumn{3}{c}{ClinTox} 
\\
\cmidrule(lr){2-4}
\cmidrule(lr){5-7}
\cmidrule(lr){8-10}
%\cmidrule(lr){11-13}
%\cmidrule(lr){14-16}

&10-shot
&5-shot
&1-shot

&10-shot
&5-shot
&1-shot

&10-shot
&5-shot
&1-shot

% &10-shot
% &5-shot
% &1-shot

% &10-shot
% &5-shot
% &1-shot
\\ 
\midrule
MAML
% &79.59$\pm$0.33 
% &77.12$\pm$0.61
% &75.63$\pm$0.18 

% &70.49$\pm$0.54 
% &70.05$\pm$0.87
% &68.63$\pm$1.51

% &68.38$\pm$1.27
% &66.47$\pm$0.32
% &65.82$\pm$2.49 

&68.40$\pm$1.02
&67.55$\pm$0.65
&61.11$\pm$1.36

&66.22$\pm$1.31
&65.25$\pm$0.75
&62.04$\pm$1.73

&74.09$\pm$2.06
&72.34$\pm$1.81
&71.61$\pm$1.24

\\
Sharp-MAML
% &75.37$\pm$0.23  
% &75.09$\pm$0.72
% &74.59$\pm$0.56 

% &71.02$\pm$ 0.81  
% &70.49$\pm$ 0.29
% &68.43$\pm$ 0.96 

% &65.52$\pm$2.01 
% &65.37$\pm$3.33
% &65.12$\pm$2.98 

&66.95$\pm$1.42
&65.41$\pm$0.95
&63.76$\pm$1.06

&73.19$\pm$2.37 
&71.25$\pm$1.84
&69.23$\pm$1.36

&77.52$\pm$2.74
&76.13$\pm$2.66
&74.47$\pm$1.81
\\

ProtoNet
% &72.99$\pm$0.56 
% &72.04$\pm$2.13
% &68.22$\pm$0.46

% &61.34$\pm$1.08 
% &60.88$\pm$1.84
% &57.41$\pm$0.76 

% &68.92$\pm$1.64
% &64.86$\pm$2.31
% &64.81$\pm$1.95

&68.87$\pm$0.82
&66.25$\pm$1.04
&58.55$\pm$1.18

&64.93$\pm$1.94
&62.29$\pm$2.12
&55.79$\pm$1.45

&76.83$\pm$1.81
&76.11$\pm$1.28
&75.42$\pm$0.86

\\
EGNN
% &80.11$\pm$0.31
% &76.80$\pm$2.62
% &75.71$\pm$0.21

% &71.24$\pm$0.37
% &69.14$\pm$1.22
% &66.36$\pm$0.29

% &68.84$\pm$1.35
% &66.92$\pm$1.27
% &62.72$\pm$1.97 

&66.42$\pm$0.77
&65.01 $\pm$0.94
&63.98 $\pm$1.20

&69.92 $\pm$1.85 
&68.35 $\pm$1.52
&67.71$\pm$ 3.67

&74.46$\pm$2.84
&74.08$\pm$3.17
&73.24$\pm$1.66

\\

Meta-MGNN

% &83.44$\pm$0.14
% &83.03$\pm$0.25
% &82.67 $\pm$0.20

% &77.84 $\pm$0.34
% &76.19$\pm$0.65
% &74.62$\pm$0.41

% &68.31 $\pm$3.06
% &67.91$\pm$ 2.09
% &66.10$\pm$ 3.98

&76.27$\pm$0.57
&75.26$\pm$0.93
&72.43$\pm$0.85

&72.58$\pm$0.34
&72.55$\pm$0.19
&72.51$\pm$0.52

&79.08$\pm$2.74
&77.81$\pm$1.64
&76.29$\pm$2.73
\\
 
\midrule

PAR
% &82.13$\pm$0.26 
% &81.92$\pm$0.37
% &80.02$\pm$0.30 

% &75.15$\pm$0.35 
% &74.01$\pm$0.12
% &72.33$\pm$0.47

% &68.08$\pm$2.23 
% &67.37$\pm$1.32
% &65.62$\pm$3.49 

&74.77$\pm$0.85
&71.25$\pm$1.17
&69.45$\pm$1.34

&78.35$\pm$2.03
&76.15$\pm$1.19
&74.02$\pm$1.68

&85.41$\pm$2.04
&85.12$\pm$1.67
&82.68$\pm$1.15

\\
TPN
% &76.05$\pm$0.24
% &75.45$\pm$0.95
% &60.16$\pm$1.18

% &67.84$\pm$0.95
% &66.52$\pm$1.28
% &62.90$\pm$1.38

% &65.22$\pm$5.82
% &65.13$\pm$0.23
% &50.00$\pm$0.51

&69.47$\pm$0.71
&66.04$\pm$1.14
&63.78$\pm$0.05

&67.61$\pm$0.33
&63.66$\pm$1.64
&60.28$\pm$1.39

&75.36$\pm$3.09
&74.25$\pm$4.02
&74.02$\pm$3.27
\\ 

GS-Meta
% &86.38$\pm$0.69 
% &84.13$\pm$0.72
% &79.32$\pm$0.89 

% &83.72$\pm$0.54
% &83.22$\pm$0.51
% &82.84$\pm$0.67 

% &67.11$\pm$1.95 
% &64.50$\pm$0.20
% &64.70$\pm$2.88

&72.58$\pm$0.34
&72.55$\pm$0.19
&72.51$\pm$0.52

&79.08$\pm$2.74 
&77.81$\pm$1.64 
&76.29$\pm$2.73

&81.02$\pm$2.12
&80.78$\pm$1.89
&80.03$\pm$4.52
\\

HSL-RG
&71.34$\pm$1.39
&70.57$\pm$0.32 
&69.81$\pm$0.66

&80.21$\pm$0.97
&\underline{79.07$\pm$0.85}
&\underline{76.69$\pm$0.38}

& 79.22$\pm$1.63
& 77.03$\pm$1.07
&76.61$\pm$0.85

\\
Pin-tuning
% &84.94$\pm$0.35
% &84.89$\pm$1.08
% &80.76$\pm$0.43

% &89.58$\pm$0.58
% &83.15$\pm$0.24
% &81.63$\pm$0.81

% &72.49$\pm$0.21
% &71.78$\pm$0.24
% &63.54$\pm$0.77

&\underline{84.75$\pm$1.41}
&\underline{83.03$\pm$1.05}
&\underline{76.98$\pm$1.92}

&\underline{80.52$\pm$0.59}
&78.35$\pm$0.64
&75.98$\pm$0.83

&\underline{89.08$\pm$2.81}
&\underline{87.63$\pm$1.44}
&\underline{83.25$\pm$4.16}

\\
\midrule

% CaMol w/o M.
% &86.95$\pm$2.58  
% &86.39$\pm$0.46
% &79.43$\pm$2.71

% &84.34$\pm$0.49  
% &83.49$\pm$0.62
% &77.06$\pm$0.31

% &91.74$\pm$2.89  
% &91.09$\pm$0.51
% &87.48$\pm$1.95
% \\

% CaMol w/o C.G.
% &81.95$\pm$0.54  
% &79.01$\pm$0.42
% &76.37$\pm$0.73

% &77.91$\pm$1.38  
% &76.21$\pm$0.36
% &74.43$\pm$0.42

% &87.89$\pm$2.05  
% &87.21$\pm$0.64
% &82.71$\pm$0.66
% \\

CaMol (Ours)

&\textbf{89.64$\pm$0.25}
&\textbf{87.46$\pm$0.29}
&\textbf{82.65$\pm$0.68}

&\textbf{87.74$\pm$0.59}
&\textbf{84.92$\pm$0.24}
&\textbf{78.26$\pm$0.41}

&\textbf{93.83$\pm$0.85}
&\textbf{92.76$\pm$0.38}
&\textbf{89.63$\pm$0.57}
\\
\bottomrule
\end{tabular}
\end{adjustbox}
\label{tab:kshot_2}
\end{table*}

\begin{table*}[t]
\caption{A model explainability comparison on functional group detection and few-shot MPP tasks in terms of Fidelity ($Fid-$/$+$).  }
\centering
\setlength{\tabcolsep}{0.5 mm } %Default value: 6pt
\fontsize{9 pt}{9 pt} \selectfont
\begin{adjustbox}{width= \linewidth}
\begin{tabular}{l cc cc cc cc cc cc}
\toprule
\multirow{2}{*}{Methods}
&\multicolumn{2}{c}{BENZENE} 
&\multicolumn{2}{c}{Alkane Carbonyl} 
&\multicolumn{2}{c}{Fluoride Carbonyl} 
&\multicolumn{2}{c}{SR-HSE} 
&\multicolumn{2}{c}{SR-MMP} 
&\multicolumn{2}{c}{SR-p53} 
\\
\cmidrule(lr){2-3}
\cmidrule(lr){4-5}
\cmidrule(lr){6-7}
\cmidrule(lr){8-9}
\cmidrule(lr){10-11}
\cmidrule(lr){12-13}
&$Fid-\downarrow$ &$Fid+\uparrow$
&$Fid-\downarrow$ &$Fid+\uparrow$
&$Fid-\downarrow$ &$Fid+\uparrow$
&$Fid-\downarrow$ &$Fid+\uparrow$
&$Fid-\downarrow$ &$Fid+\uparrow$
&$Fid-\downarrow$ &$Fid+\uparrow$
\\
\midrule

MAML 
&0.322$\pm$0.041 &0.625$\pm$0.059
&0.329$\pm$0.082 &0.561$\pm$0.057
&0.373$\pm$0.060 &0.571$\pm$0.062
&0.364$\pm$0.020 &0.545$\pm$0.081
&0.331$\pm$0.007 &0.589$\pm$0.098 
&0.386$\pm$0.003 &0.562$\pm$0.054 
\\

Sharp-MAML
&0.348$\pm$0.063 &0.585$\pm$0.105
&0.318$\pm$0.066 &0.594$\pm$0.093
&0.415$\pm$0.031 &0.606$\pm$0.044
&0.382$\pm$0.070 &0.592$\pm$0.030 
&\underline{0.260$\pm$0.085} &0.605$\pm$0.094
&0.370$\pm$0.017 &0.551$\pm$0.039
\\

ProtoNet  
&0.437$\pm$0.075 &0.625$\pm$0.091
&0.487$\pm$0.054 &0.634$\pm$0.090
&0.281$\pm$0.049 &0.548$\pm$0.071
&0.330$\pm$0.066 &0.531$\pm$0.021
&0.358$\pm$0.090 &0.608$\pm$0.003
&0.380$\pm$0.075 &0.698$\pm$0.021
\\

EGNN
&0.388$\pm$0.050 &0.616$\pm$0.072
&0.397$\pm$0.058 &0.520$\pm$0.095
&0.318$\pm$0.053 &0.681$\pm$0.074
&0.405$\pm$0.076 &0.676$\pm$0.050
&0.367$\pm$0.032 &0.616$\pm$0.081
&0.311$\pm$0.014 &0.654$\pm$0.069
\\

Meta-MGNN 
&0.329$\pm$0.048 &0.672$\pm$0.086
&0.365$\pm$0.051 &0.571$\pm$0.095
&0.446$\pm$0.049 &0.672$\pm$0.076
&0.420$\pm$0.056 &0.475$\pm$0.088
&0.387$\pm$0.068 &0.639$\pm$0.047
&0.325$\pm$0.016 &0.620$\pm$0.063
\\

\midrule
PAR   
&0.241$\pm$0.071 &\underline{0.682$\pm$0.051}
&0.415$\pm$0.054 &0.681$\pm$0.072
&0.239$\pm$0.063 &0.681$\pm$0.077
&0.378$\pm$0.043 &0.378$\pm$0.043
&0.312$\pm$0.006 &0.659$\pm$0.025
&0.374$\pm$0.004 &0.629$\pm$0.067 
\\

GS-Meta  
&\underline{0.201$\pm$0.039} &0.667$\pm$0.048
&\underline{0.234$\pm$0.050} &{0.709$\pm$0.094}
&0.245$\pm$0.038 &\underline{0.706$\pm$0.059}
&{0.306$\pm$0.060} &0.616$\pm$0.045
&0.279$\pm$0.095 &0.655$\pm$0.021
&\underline{0.292$\pm$0.049} &\textbf{0.719$\pm$0.035}
\\

TPN 
&0.312$\pm$0.071 &0.591$\pm$0.065
&0.274$\pm$0.048 &0.685$\pm$0.063
&0.248$\pm$0.085 &0.662$\pm$0.078
&0.322$\pm$0.025 &0.625$\pm$0.033 
&0.385$\pm$0.053 &0.573$\pm$0.038 
&0.322$\pm$0.062 &0.599$\pm$0.015
\\

Pin-Tuning  
&0.226$\pm$0.041 &0.628$\pm$0.072
&0.257$\pm$0.073 &\underline{0.719$\pm$0.093}
&\underline{0.234$\pm$0.031} &0.689$\pm$0.072
&\underline{0.306$\pm$0.027} &\underline{0.683$\pm$0.031} 
&0.340$\pm$0.064 &\underline{0.695$\pm$0.096}
&0.308$\pm$0.055 &0.667$\pm$0.085
\\ 

\midrule
CaMol (Ours)
&\textbf{0.128$\pm$0.013} &\textbf{0.741$\pm$0.060}
&\textbf{0.158$\pm$0.022} &\textbf{0.755$\pm$0.052}
&\textbf{0.144$\pm$0.013} &\textbf{0.762$\pm$0.051}
&\textbf{0.264$\pm$0.008} &\textbf{0.697$\pm$0.063}
&\textbf{0.207$\pm$0.023} &\textbf{0.701$\pm$0.075}
&\textbf{0.203$\pm$0.052} &\underline{0.713$\pm$0.088}
\\
\bottomrule
\end{tabular}
\end{adjustbox}
\label{tab:fidelity_1}
\end{table*}

\begin{table}[tb]
\centering
\fontsize{9 pt}{9 pt} \selectfont
\setlength{\tabcolsep}{1  mm } %Default value: 6pt
\caption{
A consistency comparison for causal substructures discovered from each property in terms of JSD scores.  
(BEN.: BENZENE, A.C.: Alkane Carbonyl, F.C.: Fluoride Carbonyl)
}
\begin{adjustbox}{width= \linewidth}
\begin{tabular}{l ccc ccc}
\toprule
\multirow{2}{*}{Methods}
&\multicolumn{3}{c}{Explainable datasets} 
&\multicolumn{3}{c}{Tox21 dataset} 
\\
\cmidrule(lr){2-4}
\cmidrule(lr){5-7}

&BEN.
&A.C.
&F.C.
%&Alkane Carbonyl
%&Fluoride Carbonyl

&SR-HSE 
&SR-MMP 
&SR-p53
\\
\midrule

MAML
&0.440 &0.452 &0.502
&0.575 &0.492 &0.470
\\
Sharp-MAML
&0.416 &0.419 &0.478
&0.444 &0.447 &0.437
\\
ProtoNet
&0.423 &0.431 &0.462
&0.436 &0.481 &0.369
\\
EGNN
&0.413 &0.497 &0.437
&0.486 &0.511 &0.391
\\
Meta-MGNN
&0.372 &0.392 &0.503
&0.578 &0.407 &0.343
\\
\midrule
PAR
&0.360 &0.492 &0.519
&0.492 &0.353 &0.461
\\
GS-Meta
&0.318 &\underline{0.392} &0.390
&\underline{0.308} &0.344 &0.302
\\
TPN
&0.456 &0.503 &0.494
&0.411 &0.398 &0.326
\\
HSL-RG
&0.452 &0.684 &0.428
&0.362 &0.331 &0.356
\\
Pin-Tuning
&\underline{0.303} &0.414 &\underline{0.386}
&0.368 &\underline{0.287} &\underline{0.292}
\\
\midrule
CaMol (Ours)
&\textbf{0.205} &\textbf{0.252} &\textbf{0.311}
&\textbf{0.205} &\textbf{0.189} &\textbf{0.150}
\\

\bottomrule
\end{tabular}
\end{adjustbox}
\label{tab:jsd_tox21}
\end{table}

\subsection{Experimental Settings}

\subsubsection{Datasets}

We evaluate CaMol on six widely used few-shot molecular property prediction datasets from MoleculeNet \cite{wu2018moleculenet}, including Tox21, SIDER, MUV, ToxCast, PCBA, and ClinTox.
For the model explainability, we used three datasets with ground truth substructure explanations, i.e., Benzene, Alkane Carbonyl, and Fluoride Carbonyl \cite{agarwal2023evaluating}.
We follow the standard data splits commonly adopted in Few-shot MPP to ensure fair comparison with prior work \cite{doi:10.1021/acscentsci.6b00367}.
The dataset statistics are given in Appendix C. %\ref{appendix:dataset}.

\subsubsection{Baselines and Implementation Details}

%For a comprehensive comparison, 
We adopt two groups of baselines.
(1) Traditional methods include:
MAML \cite{DBLP:conf/icml/FinnAL17},
Sharp-MAML \cite{DBLP:conf/icml/AbbasXCCC22},
ProtoNet \cite{DBLP:conf/nips/SnellSZ17},
EGNN \cite{DBLP:conf/cvpr/KimKKY19}, and
Meta-MGNN \cite{DBLP:conf/www/GuoZYHW0C21}.
(2) In-context learning methods include:
PAR \cite{DBLP:conf/nips/WangAYD21},
GS-Meta \cite{DBLP:conf/ijcai/ZhuangZWDFC23},
TPN \cite{DBLP:conf/ijcai/MaBALLZL20},
HSL-RG \cite{DBLP:journals/nn/JuLQFWG0023}, and
Pin-Tuning \cite{DBLP:conf/nips/0006LSW024}.
For the pre-trained encoder, we adopt S-CGIB \cite{Hoang_Lee_2025} as the backbone encoder.
We employ a 3-layer EGIN \cite{DBLP:conf/iclr/HuLGZLPL20} as the context graph encoder.
The implementation details and reproducibility are provided in Appendix D.
We deliver an open-source implementation of CaMol for the experiment reproductions\footnote{https://github.com/NSLab-CUK/CaMol}.
%\ref{app:imp_details}.

%\subsubsection{Evaluation Metrics}
%\subsubsection{Implementation Details}

\subsection{Few-shot Performance Analysis}

We conducted few-shot experiments on six molecular property prediction datasets under various few-shot settings, as shown in Table~\ref{tab:kshot_2}.
We observed that:
(1) CaMol consistently achieved the best performance across all settings, with an average relative improvement of 7.36\% over the strongest in-context learning baseline.
For example, on Tox21 (10-shot), CaMol reached 92.69\% ROC-AUC, outperforming GS-Meta (86.38\%) by 6.31\%.
These results indicate that causal substructure discovery and functional group reasoning provide strong inductive biases for few-shot molecular property prediction.
(2) In-context learning methods, e.g., Pin-Tuning, PAR, and GS-Meta, outperformed traditional meta-learning approaches.
This indicates the effectiveness of leveraging molecular–property relationships in addressing the few-shot tasks.
(3) CaMol showed large gains on datasets with high structural diversity and class imbalance, such as MUV and PCBA.
For example, on MUV (1-shot), CaMol achieved 76.79\%, compared to 65.62\% for PAR, a relative improvement of 17.06\%.
By exploiting chemical signals, CaMol effectively discovered transferable causal substructures, showing strong generalization to unseen properties and high performance across benchmarks.

\begin{figure*}[tb]
\centering
  \includegraphics[width=  \linewidth]{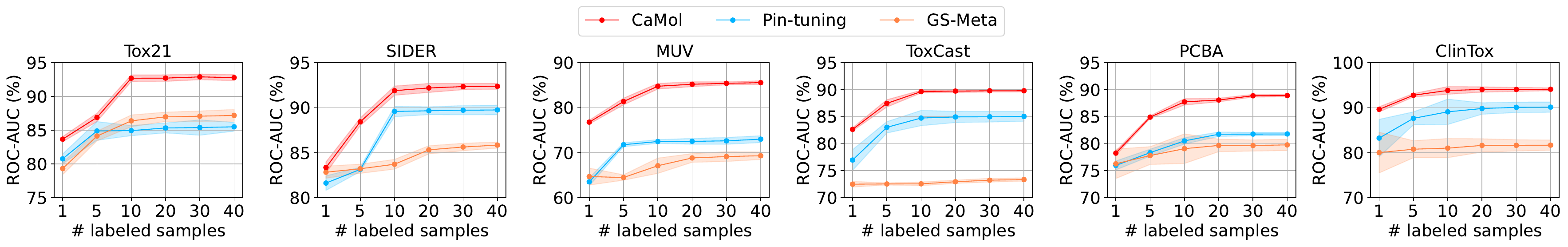}
  \caption{A sample efficiency comparison on Few-shot MPP in terms of ROC-AUC according to the number of labeled samples.}
  \label{fig:ab_efficiency}
\end{figure*}

\begin{figure}[tb]
\centering
  \includegraphics[width=  \linewidth]{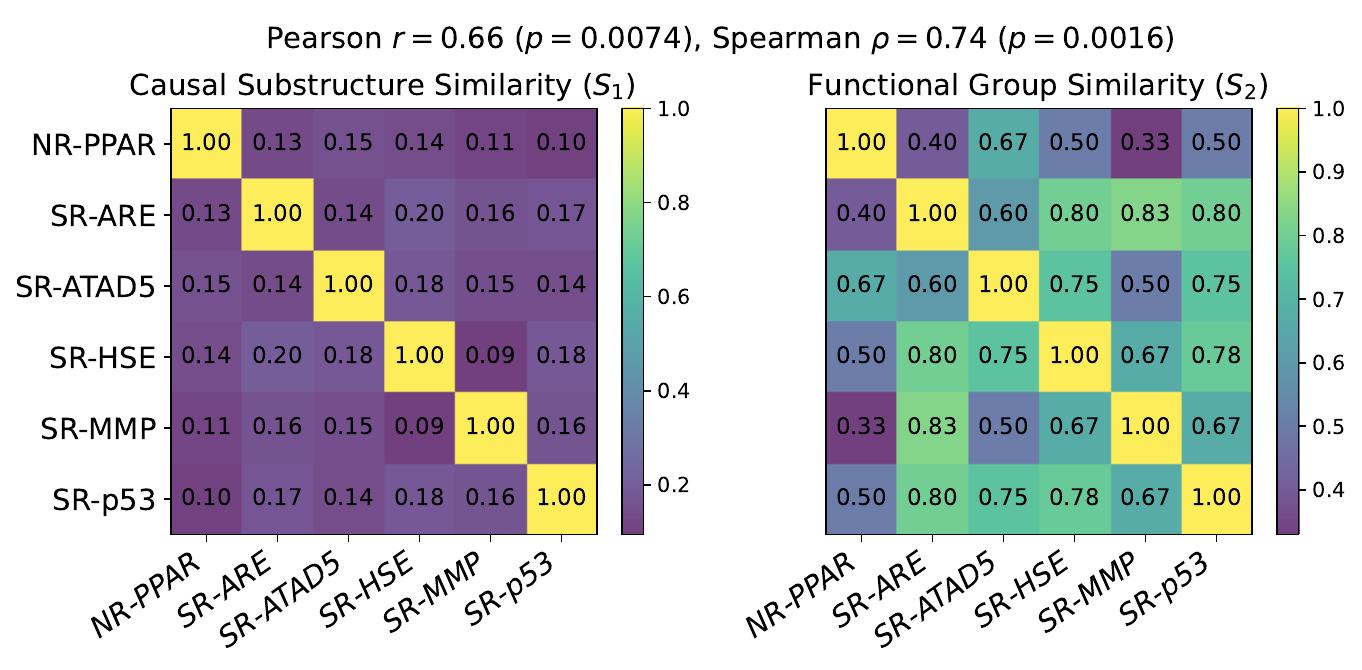}
% \caption{Heatmap of property–property similarity relations across six properties in Tox21 dataset. 
% The first shows similarities based on found causal subgraph embeddings (inverse JSD scores), the second shows similarities based on functional group frequencies, and the third shows their normalized absolute difference.
% Strong agreement was observed between the two approaches (Pearson $r = 0.67$, Spearman $\rho = 0.74$; both $p < 0.01$).}
% \caption{Heatmaps of property–property similarity relations obtained from (left) causal subgraph embeddings, (middle) functional group frequencies, and (right) their absolute difference. 
% Strong agreement is observed between the two strategies (Pearson $r = 0.67$, 95\% CI [0.61, 0.71]; Spearman $\rho = 0.74$, 95\% CI [0.70, 0.78]; both $p < 0.01$).}
\caption{Heatmaps of property–property similarities obtained from (left) representations of causal substructure discovered by CaMol and (right) co-occurrence frequencies of functional groups.
The two similarities are highly correlated (Pearson $r = 0.67$, Spearman $\rho = 0.74$; both $p < 0.01$).
%Strong agreement is observed (Pearson $r = 0.67$, 95\% CI [0.61, 0.71]; Spearman $\rho = 0.74$, 95\% CI [0.70, 0.78]; both $p < 0.01$).
}
\label{fig:heat_map}
\end{figure}

\subsection{Molecule Interpretation Analysis}
We now evaluate whether the found causal substructures remain consistent across molecules for predicting similar properties, as shown in Table~\ref{tab:jsd_tox21}.
We use three explainable datasets with ground-truth substructure annotations and three properties from the Tox21 dataset.
Specifically, we select the top 50\% of nodes with the highest probability scores from CaMol’s masking strategy as explainable nodes.
For baselines, we identify the top 50\% of nodes with the highest positive saliency values as explainable substructures, following \citet{DBLP:conf/cvpr/PopeKRMH19}.
To measure consistency, we compute the Jensen–Shannon Divergence (JSD) between the found causal substructure embeddings with similar properties.
We observed that:
{(1)} 
CaMol consistently outperformed baselines that do not discover causal substructures during training, e.g., Pin-tuning, GS-Meta, and Meta-MGNN, across explainable datasets.
For example, on the BENZENE dataset, CaMol reduced JSD by 32.3\% relative to the second-best model, showing that its discovered substructures are more consistent and property-relevant.
This implies that CaMol could learn the correct and stable substructures that respond to the given properties.
{(2)}
The found causal substructures could also improve prediction on unseen properties in Tox21.
For example, on the SR-HSE property, CaMol achieved a 33.4\% improvement over baselines, emphasizing that causal substructure discovery provides transferable inductive bias for few-shot generalization.
To sum up, CaMol could identify minimal yet sufficient substructures that are consistent and interpretable to similar properties.

Since we use functional groups as prior knowledge to guide the causal substructure discovery, we validate whether the found causal substructures preserve chemical knowledge about functional group relationships.
To achieve it, we conducted a heatmap analysis of property–property similarity on six properties in the Tox21, as shown in Figure~\ref{fig:heat_map}.
Specifically, we compared the similarity of substructures derived from CaMol’s causal substructure embeddings with those obtained directly from functional group frequencies.
The causal embeddings are then aggregated across molecules for each property, and pairwise similarities are computed using the inverse JSD score, while functional group–based similarities are calculated using the Jaccard score.
We observed a strong agreement between the two similarity matrices, with Pearson $r = 0.67$, 95\% Confidence Interval [0.61, 0.71] and Spearman $\rho = 0.74$, 95\% Confidence Interval [0.70, 0.78], both $p < 0.01$.
The heatmap revealed that properties sharing common functional groups cluster together in both spaces, while chemically distinct properties remain well separated.
The finding indicates that CaMol effectively captured and preserved prior knowledge even in the few-shot scenarios.

%To further demonstrate the global consistency of the found causal substructures with chemical knowledge, we conducted a heatmap analysis of property–property similarity relations on six properties in the Tox21 dataset, as shown in Figure~\ref{fig:heat_map}.
% In this analysis, we compare the similarity relations derived from the causal substructure embeddings discovered by CaMol with those obtained directly from functional group frequencies.
% The causal substructure embeddings are aggregated across molecules for each property, and pairwise similarities between properties are computed using the inverse JSD score.
% For functional group–based similarity, we calculate the Jaccard similarity between properties that share common functional groups across molecules.
% This evaluation enables us to validate whether the found causal substructures capture relationships aligned with real-world functional group distributions, even under the few-shot prediction of unseen properties.
% We observed a strong agreement between the two similarity matrices, with Pearson $r = 0.67$ and Spearman $\rho = 0.74$, both $p < 0.01$.
% This demonstrates that CaMol is able to recover causal substructures that are not only predictive for molecular properties but also consistent with domain knowledge about functional group–property relationships.
% This finding validates CaMol’s ability to generalize to unseen properties in few-shot scenarios by capturing the correct substructures across molecules.

\subsection{Model Explainability Analysis}

%There has been concern regarding the explainability of GNNs, which are black-box.
%Beyond achieving strong predictive performance, CaMol also provides interpretable explanations of its predictions.
To evaluate CaMol’s explainability, we compared its discovered explanations against ground-truth structural annotations~\cite{agarwal2023evaluating}.
We extracted explanations by selecting the top 50\% of nodes with the highest probability $p$, emphasizing the causal substructures most responsible for the prediction.
We then used the \textit{Fidelity+/-} metrics~\cite{amara2022graphframex} to assess how well these explanations aligned with the model’s own predictions, as reported in Table~\ref{tab:fidelity_1}.
For the baselines, following~\citet{DBLP:conf/cvpr/PopeKRMH19}, we selected the top 50\% of nodes with the highest positive saliency values as explanations.
We observed that CaMol consistently achieved the best explanation quality on both metrics across all datasets.
For example, on BENZENE, CaMol reduced $Fidelity-$ by 36.6\% compared to the strongest baseline, showing that its discovered substructures aligned more closely with the model’s own predictions.
This indicated that CaMol not only improved prediction accuracy but also enhanced interpretability by providing faithful explanations.

% To evaluate CaMol’s explainability, we compare its discovered explanations against ground-truth structural annotations~\cite{agarwal2023evaluating}.
% We extract explanations by selecting the top 50\% of nodes with the highest probability $p$, which emphasize the causal substructures most responsible for the prediction.
% We then use the \textit{Fidelity+/-} metrics~\cite{amara2022graphframex} to assess how well these explanations align with the model’s own predictions, as reported in Table~\ref{tab:fidelity_1}.
% For baselines, we follow~\citet{DBLP:conf/cvpr/PopeKRMH19} and select the top 50\% of nodes with the highest positive saliency values as explanation.
% We observed that CaMol consistently achieved the best explanation quality on both metrics across all datasets.
% For example, on BENZENE, CaMol reduces $Fidelity-$ by 36.6\% compared to the strongest baseline, showing that its discovered substructures align more closely with the model’s own predictions.
% This indicates that CaMol not only improves prediction accuracy but is also aligned with the explanations.

\begin{table}[tb]
\centering
\setlength{\tabcolsep}{1 mm } %Default value: 6pt
\fontsize{9 pt}{9 pt}\selectfont
\caption{Ablation study on different components of CaMol,
intervention strategies,
and loss functions.}
\label{tab:ab_Camol_12}

\begin{tabular}{l ccc ccc}
\toprule
& Tox21 & SIDER & MUV & ToxCast & PCBA & ClinTox \\
\midrule

\multicolumn{7}{c}{\textbf{(a) On different components of CaMol.}} \\ 
\midrule
w/o C.G. 
&78.66 &77.7 &75.18 &79.01 &76.21 &87.21 \\

w/o F.G.
&\underline{85.11} &\underline{86.59} &80.14 &\underline{87.11} &\underline{84.45} &90.35 \\

w/o Causality
&84.95 &86.15 &\underline{80.22} &86.39 &83.49 &\underline{91.09} \\

CaMol
&\textbf{86.89} &\textbf{88.43} &\textbf{81.37} &\textbf{87.46} &\textbf{84.92} &\textbf{92.76} \\

\midrule
\multicolumn{7}{c}{\textbf{(b) On different intervention strategies.}} \\ 
\midrule

Aug-Int
&85.11 &87.81 &80.95 &86.61 &\underline{84.41} &92.51 \\

Rand-Int
&\underline{86.19} &\underline{87.94} &\underline{81.02} &\underline{87.14} &84.22 &\underline{92.54} \\

w/o-Int
&85.01 &87.13 &79.94 &87.03 &83.66 &91.31 \\

FG-Int (Ours)
&\textbf{86.89} &\textbf{88.43} &\textbf{81.37} &\textbf{87.46} &\textbf{84.92} &\textbf{92.76} \\

\midrule
\multicolumn{7}{c}{\textbf{(c) On the contribution of different loss functions.}} \\ 
\midrule

w/o $\mathcal{L}_{KL}$
&85.01 &\underline{87.83} &79.94 &\underline{87.03} &\underline{83.66} &\underline{91.31} \\

w/o $\mathcal{L}_{var}$
&\underline{86.02} &87.14 &\underline{80.78} &85.66 &82.14 &89.35 \\

w/o $\mathcal{L}_{causal}$
&84.95 &86.15 &80.22 &86.39 &83.49 &91.09 \\

CaMol
&\textbf{86.89} &\textbf{88.43} &\textbf{81.37} &\textbf{87.46} &\textbf{84.92} &\textbf{92.76} \\

\bottomrule
\end{tabular}
\end{table}

\subsection{Model Efficiency Analysis}

We evaluated the generalization and sample efficiency of CaMol across different numbers of labeled molecules, as shown in Figure~\ref{fig:ab_efficiency}.
CaMol consistently demonstrated strong generalization ability and high stability in few-shot scenarios, showing competitive performance even with extremely limited supervision.
For example, with only 10 labeled molecules in the Tox21 dataset, CaMol achieved stable high accuracy, showing only marginal improvement as more samples were added.
This finding indicated that CaMol effectively learned transferable causal representations that generalized well to unseen properties under data-scarce conditions.
We further conducted an efficiency analysis on the trade-off between performance and parameter size (Appendix E.1). % \ref{app:exp_trade_off}).
We observed that CaMol achieved the best accuracy–efficiency trade-off, positioned on the high–left Pareto frontier.
%, surpassing both in-context and traditional methods.
%Overall, this suggested that using causal substructures and contextual signals not only enhanced interpretability but also significantly improved sample efficiency and generalization in few-shot MPP.

% We evaluated the sample efficiency of CaMol across different shot scenarios, as shown in Figure~\ref{fig:ab_efficiency}.
% We observed that CaMol showed higher sample efficiency and exhibited lower variability when trained on a few labeled molecules.
% For example, with only 10 labeled molecules in the Tox21 dataset, CaMol could achieve stable high performance, and its accuracy remains consistent as more labeled samples are added.
% This result indicates that CaMol can effectively learn meaningful causal patterns from only a few samples, enabling robust performance even in extremely data-scarce settings.
% %We attribute this stability to CaMol’s ability to focus on causal substructures, which provide the correct inductive bias for generalization in few-shot MPP.
% We further conducted an efficiency analysis on the trade-off between performance and parameter size (Appendix \ref{app:exp_trade_off}).
% We observed that CaMol achieves the best accuracy–efficiency trade-off, positioned on the high–left Pareto frontier, surpassing both in-context and traditional methods.

%\subsubsection{Sample Efficiency}
%\subsubsection{Parameter Efficiency}

\subsection{Ablation Analysis}

%\subsubsection{Ablation Study}

% Table \ref{tab:ab_components}: Ablation study on different components of CaMol: w/o Context Graph, w/o Functional Group, and w/o Masking.

\subsubsection{Model Components}

To evaluate the contribution of each component in CaMol, we remove the context graph (w/o C.G.), functional groups (w/o F.G.), and the masking strategy (w/o Causality), as shown in Table~\ref{tab:ab_Camol_12} (a).
We observed that:
(1) Removing the context graph results in the largest performance drop across all datasets.
For example, on Tox21, performance decreased from 86.89 to 78.66 ROC-AUC, showing that relational signals among molecules, functional groups, and properties are essential for discovering causal substructures.
(2) Removing functional groups within the context graph also leads to degradation in performance, e.g., from 88.43 to 86.59 on the SIDER dataset. This implies that functional groups provide an important inductive bias by serving as chemical prior knowledge, which is valuable in few-shot settings.
(3) Removing the masking mechanism reduces performance moderately, e.g., from 81.37 to 80.22 on the MUV dataset.
Although the impact is smaller compared to the context graph, the consistent decline across datasets implies that masking is important to discover causal substructures, improving prediction on unseen properties.
Overall, the context graph delivered relational signals, while graph causal inference focused on key substructures, jointly enabling CaMol’s robust and generalizable performance.
%For example, on the SIDER dataset, ROC-AUC drops from 88.43 to 86.59, confirming their role as informative inductive biases in few-shot scenarios.
% Overall, the context graph provides relational guidance, functional groups act as domain-informed causal priors, and the masking strategy enforces causal feature selection.
% Together, these components enable CaMol to achieve robust performance across diverse molecular property prediction tasks.

%Table \ref{tab:ab_intervener}: Ablation study on different intervention strategies. (FG-Int: functional group intervention; Aug-Int: augmentation intervention; Rand-Int: random noise intervention).

\subsubsection{Intervention Strategies}

%We further investigate the effect of different intervention strategies, as shown in Table~\ref{tab:ab_Camol_12} (b).
%We compared four variants: functional group intervention (FG-Int, our proposed strategy), augmentation-based intervention (Aug-Int), random noise intervention (Rand-Int), and the model without intervention (w/o-Int).
We evaluated four intervention strategies: functional group–based (FG-Int, our method), augmentation-based (Aug-Int), random noise (Rand-Int), and no intervention (w/o-Int) under the 5-shot setting, as shown in Table~\ref{tab:ab_Camol_12} (b).
We observed that FG-Int consistently achieved the best and most stable performance across datasets.
For example, on Tox21, FG-Int reached 86.89 ROC-AUC, compared to 85.11 for Aug-Int, 86.19 for Rand-Int, and 85.01 without intervention.
This implies that interventions grounded in functional groups provided semantically meaningful priors, allowing the model to disentangle causal substructures more effectively than other interveners.

% \begin{figure}[!ht]
% \centering
%   \includegraphics[width=  \linewidth]{Figures/camol_ab_study.pdf}
%   \caption{Ablation study on different Loss functions.}
%   \label{fig:ab_lossfunction}
% \end{figure}

\subsubsection{Loss Functions}

We further evaluated the contribution of different loss components in CaMol, namely $\mathcal{L}{\text{causal}}$, $\mathcal{L}{\text{KL}}$, and $\mathcal{L}{\text{var}}$, under the 5-shot setting, as shown in Table~\ref{tab:ab_Camol_12}(c).
We found that removing any of the loss terms consistently led to a decline in performance compared to CaMol.
For example, on Tox21, excluding $\mathcal{L}_{\text{causal}}$ caused the large drop, from 86.89 to 84.54 ROC-AUC, underscoring its importance in enforcing causal substructure discovery.
Removing $\mathcal{L}_{\text{KL}}$ or $\mathcal{L}_{\text{var}}$ also reduced performance, indicating that both regularization terms contributed to training stability and generalization.
Overall, these results showed that all three objectives worked complementarily to achieve CaMol’s overall performance.
% We further conduct an ablation study to evaluate the contribution of different loss components in CaMol: $\mathcal{L}_{KL}$, $\mathcal{L}_{var}$, and $\mathcal{L}_{causal}$, as shown in Table~\ref{tab:ab_Camol_12} (c).
% We observed that removing any of the loss terms consistently leads to performance degradation compared to the full CaMol model.  
% For example, on Tox21, dropping $\mathcal{L}_{causal}$ results in the largest decrease (from 86.89 to 84.54 ROC-AUC), highlighting the importance of enforcing causal substructure discovery.  
% %On SIDER, performance declines from 88.43 to 86.59 when $\mathcal{L}_{causal}$ is removed, confirming its critical role across datasets.  
% The removal of $\mathcal{L}_{KL}$ and $\mathcal{L}_{var}$ also causes noticeable drops, though to a lesser extent, indicating that both regularization terms contribute to stabilizing training and improving generalization.  
% This implies that all three losses are complementary and essential.  
% $\mathcal{L}_{causal}$ ensures that found substructures are causally meaningful, while $\mathcal{L}_{KL}$ and $\mathcal{L}_{var}$ provide regularization that prevents overfitting and implements consistency.  
% %Together, they enable CaMol to achieve state-of-the-art performance in few-shot molecular property prediction.

\subsection{Sensitivity Analysis}

%In the sensitivity analysis of subgraph size and interatomic distance, MVCIB achieved the optimal performance when the subgraph size is set to 3 and the interatomic distance threshold is 1.5~\text{\AA} (Appendix D.7).
%Regarding the impact of GNN depth, we observed that MVCIB performed well with $k=4$ layers (Appendix D.8).

We conducted sensitivity analysis by varying causal subgraph ratios and measuring the consistency and model performance under the 5-shot setting (Appendix E.2). %\ref{app:exp_sub_rarios}).
We observed that CaMol consistently yielded the lowest JSD at mid-range ratios, from 0.5 to 0.6, which balanced minimal yet sufficient causal substructures and achieved competitive accuracy.
Small ratios could discard key atoms, while larger ratios introduced noise, which confirmed CaMol’s robustness in capturing minimal yet consistent causal substructures.

We further analyzed the effect of the hyperparameters $\alpha_1$ and $\alpha_2$, which controlled $\mathcal{L}_{\text{KL}}$ and $\mathcal{L}_{\text{var}}$, respectively under the 5-shot setting (Appendix E.3). %\ref{app:sen_1}).
We observed that CaMol achieved the best performance with moderate values of $\alpha_1$ and $\alpha_2$.
That is, the small values weakened the disentanglement and intervention effects, while too large values over-regularized the model.
Regarding GNN depth, we observed that CaMol performed best at $k=3$ layers, as shown in Appendix E.4. %\ref{app:sen_2}.

\section{Conclusion}

In this paper, we propose CaMol, a novel context-aware causal inference framework for few-shot molecular property prediction.
CaMol leverages contextual signals among functional groups and properties to guide causal substructure discovery under the graph causality inference.
By integrating a context graph, a learnable masking mechanism, and the chemically grounded interventions, CaMol separates causal substructures from confounding variations. 
Extensive experiments demonstrated that CaMol consistently outperformed baselines, achieving superior accuracy and sample efficiency in few-shot settings.
Moreover, the found causal substructures showed strong alignment with chemical knowledge of functional groups, providing interpretable insights into molecular behavior.

\bibliography{aaai25}

@String{Computing = "Computing" }

@String{Computer = "{IEEE} Computer" }

@String{Chelsea = "Chelsea" }

@BOOK{test,
   author = "Donald E. Knuth",
   title = "Seminumerical Algorithms",
   volume = 2,
   series = "The Art of Computer Programming",
   publisher = "Addison-Wesley",
   address = "Reading, MA",
   edition = "2nd",
   month = "10~" # jan,
   year = "1981",
}

@ArtifactSoftware{R,
    title = {R: A Language and Environment for Statistical Computing},
    author = {{R Core Team}},
    organization = {R Foundation for Statistical Computing},
    address = {Vienna, Austria},
    year = {2019},
    url = {https://www.R-project.org/},
}

@InProceedings{DBLP:conf/iclr/XuHLJ19,
  author    = {Keyulu Xu and Weihua Hu and Jure Leskovec and Stefanie Jegelka},
  booktitle = {Proceedings of the 7th International Conference on Learning Representations ({ICLR} 2019)},
  title     = {How Powerful are Graph Neural Networks?},
  year      = {2019},
  address   = {New Orleans, LA, USA},
  month     = {May},
  publisher = {OpenReview.net},
  bibsource = {dblp computer science bibliography, https://dblp.org},
}

@inproceedings{DBLP:conf/iclr/HuLGZLPL20,
  author    = {Weihua Hu and
               Bowen Liu and
               Joseph Gomes and
               Marinka Zitnik and
               Percy Liang and
               Vijay S. Pande and
               Jure Leskovec},
  title     = {Strategies for Pre-training Graph Neural Networks},
  booktitle = {Proceedings of the 8th International Conference on Learning Representations ({ICLR} 2020)},
    address = {Addis Ababa, Ethiopia},
    month = {Apr}, 
  publisher = {OpenReview.net},
  year      = {2020},
  bibsource = {dblp computer science bibliography, https://dblp.org}
}

@article{article,
author = {Bhardwaj, Atul and Kaur, Jatinder and Wuest, Melinda and Wuest, Frank},
year = {2017},
month = {12},
pages = {},
title = {In situ click chemistry generation of cyclooxygenase-2 inhibitors},
volume = {8},
journal = {Nature Communications},
doi = {10.1038/s41467-016-0009-6}
}

@article{wang2019dgl,
    title={Deep Graph Library: A Graph-Centric, Highly-Performant Package for Graph Neural Networks},
    author={Minjie Wang and Da Zheng and Zihao Ye and Quan Gan and Mufei Li and Xiang Song and Jinjing Zhou and Chao Ma and Lingfan Yu and Yu Gai and Tianjun Xiao and Tong He and George Karypis and Jinyang Li and Zheng Zhang},
    year={2019},
    journal={arXiv preprint:1909.01315}
}

@inproceedings{
amara2022graphframex,
title={GraphFramEx: Towards Systematic Evaluation of Explainability Methods for Graph Neural Networks},
author={Kenza Amara and Zhitao Ying and Zitao Zhang and Zhichao Han and Yang Zhao and Yinan Shan and Ulrik Brandes and Sebastian Schemm and Ce Zhang},
booktitle={Proceedings of the 1st Learning on Graphs Conference (LoG 2022)},
year={2022},
address= {Virtual Event},
}

@inproceedings{DBLP:conf/cvpr/PopeKRMH19,
  author       = {Phillip E. Pope and
                  Soheil Kolouri and
                  Mohammad Rostami and
                  Charles E. Martin and
                  Heiko Hoffmann},
  title        = {Explainability Methods for Graph Convolutional Neural Networks},
  booktitle    = {Proceedings of the {IEEE} Conference on Computer Vision and Pattern Recognition ({CVPR} 2019) },
address= {Long Beach, CA, USA},
month = {June 16-20, 2019},
  pages        = {10772--10781},
  publisher    = {Computer Vision Foundation / {IEEE}},
  year         = {2019},
  bibsource    = {dblp computer science bibliography, https://dblp.org}
}

@article{agarwal2023evaluating,
  title={Evaluating explainability for graph neural networks},
  author={Agarwal Chirag and Queen Owen and Lakkaraju Himabindu and Zitnik Marinka},
  journal={Scientific Data},
  volume={10},
  number={1},
  year={2023},
  publisher={Nature Publishing Group UK London}
}

@article{degen2008art,
  title={On the art of compiling and using'drug-like'chemical fragment spaces},
  author={Degen, Jorg and Wegscheid-Gerlach, Christof and Zaliani, Andrea and Rarey, Matthias},
  journal={ChemMedChem},
  volume={3},
  number={10},
  pages={1503},
  year={2008}
}

@inproceedings{DBLP:conf/iclr/MaddisonMT17,
title={The Concrete Distribution: A Continuous Relaxation of Discrete Random Variables},
author={Chris J. Maddison and Andriy Mnih and Yee Whye Teh},
booktitle={Proceedings of the 5th International Conference on Learning Representations (ICLR 2017)},
year={2017},
address= {Toulon, France},
month = {Apr 24 - 26, 2017}, 
url={https://openreview.net/forum?id=S1jE5L5gl}
}

@inproceedings{DBLP:conf/iclr/JangGP17,
  author       = {Eric Jang and
                  Shixiang Gu and
                  Ben Poole},
  title        = {Categorical Reparameterization with Gumbel-Softmax},
  booktitle    = {Proceedings of the 5th International Conference on Learning Representations ({ICLR} 2017)},
  publisher    = {OpenReview.net},
address= {Toulon, France},
month = {Apr 24 - 26, 2017}, 
  year         = {2017},
  bibsource    = {dblp computer science bibliography, https://dblp.org}
}

@article{wu2018moleculenet,
  title={MoleculeNet: a benchmark for molecular machine learning},
  author={Wu, Zhenqin and Ramsundar, Bharath and Feinberg, Evan N and Gomes, Joseph and Geniesse, Caleb and Pappu, Aneesh S and Leswing, Karl and Pande, Vijay},
  journal={Chemical science},
  volume={9},
  number={2},
  year={2018},
  publisher={Royal Society of Chemistry}
}

@inproceedings{FeyLenssen2019,
  title={Fast Graph Representation Learning with {PyTorch Geometric}},
  author={Fey, Matthias and Lenssen, Jan E.},
  booktitle={Proceedings of the ICLR Workshop on Representation Learning on Graphs and Manifolds (ICLRW 2019)},
  year={2019},
}

@inproceedings{Hoang_Lee_2025, 
title={Pre-Training Graph Neural Networks on Molecules by Using Subgraph-Conditioned Graph Information Bottleneck}, 
volume={39}, 
number={16}, 
booktitle={Proceedings of the 39th Conference on Artificial Intelligence (AAAI 2025)}, 
author={Hoang, Van Thuy and Lee, O-Joun},
year={2025}, 
month={Apr.}, 
pages={17204-17213}
}

@inproceedings{DBLP:conf/nips/Fan0MST22,
  author       = {Shaohua Fan and
                  Xiao Wang and
                  Yanhu Mo and
                  Chuan Shi and
                  Jian Tang},
  title        = {Debiasing Graph Neural Networks via Learning Disentangled Causal Substructure},
  booktitle    = {Proceedings of the 35th Annual Conference on Neural Information Processing Systems 2022 (NeurIPS 2022)},
month = {November 28 - December 9, 2022},
address= {New Orleans,  LA, USA},
  year         = {2022},
  bibsource    = {dblp computer science bibliography, https://dblp.org}
}

@inproceedings{DBLP:conf/nips/0002ZB00XL0C22,
  author       = {Yongqiang Chen and
                  Yonggang Zhang and
                  Yatao Bian and
                  Han Yang and
                  Kaili Ma and
                  Binghui Xie and
                  Tongliang Liu and
                  Bo Han and
                  James Cheng},
  title        = {Learning Causally Invariant Representations for Out-of-Distribution
                  Generalization on Graphs},
  booktitle    = {Proceedings of the 35th Annual Conference on Neural Information Processing Systems 2022 (NeurIPS 2022)},
month = {Nov. 28 - Dec. 9, 2022},
address= {New Orleans,LA, USA},
  year         = {2022},
  bibsource    = {dblp computer science bibliography, https://dblp.org}
}

@inproceedings{DBLP:conf/kdd/SuiWWL0C22,
  author       = {Yongduo Sui and
                  Xiang Wang and
                  Jiancan Wu and
                  Min Lin and
                  Xiangnan He and
                  Tat{-}Seng Chua},
  title        = {Causal Attention for Interpretable and Generalizable Graph Classification},
  booktitle    = {Proceedings of the 28th Conference on Knowledge Discovery
                  and Data Mining (KDD 2022)},
month = {August 14 - 18, 2022},
address= {Washington, DC, USA},
  pages        = {1696--1705},
  publisher    = {{ACM}},
  year         = {2022},
  bibsource    = {dblp computer science bibliography, https://dblp.org}
}

@inproceedings{DBLP:conf/iclr/WuWZ0C22,
  author       = {Yingxin Wu and
                  Xiang Wang and
                  An Zhang and
                  Xiangnan He and
                  Tat{-}Seng Chua},
  title        = {Discovering Invariant Rationales for Graph Neural Networks},
booktitle    = {Proceedings of the 10th International Conference on Learning Representations (ICLR 2022)},
month = {April 25-29, 2022},
address= {Virtual},     
  publisher    = {OpenReview.net},
  year         = {2022},
  bibsource    = {dblp computer science bibliography, https://dblp.org}
}

@inproceedings{DBLP:conf/cvpr/YuLH23,
  author       = {Junchi Yu and
                  Jian Liang and
                  Ran He},
  title        = {Mind the Label Shift of Augmentation-based Graph {OOD} Generalization},
  booktitle    = {Proceedings of Conference on Computer Vision and Pattern Recognition 2023 ({CVPR} 2023) },
  pages        = {11620--11630},
address= {Vancouver, BC, Canada},
month = {June 17-24, 2023},
  publisher    = {{IEEE}},
  year         = {2023},
  bibsource    = {dblp computer science bibliography, https://dblp.org}
}

@inproceedings{DBLP:conf/nips/ZhuangZDBWLCC23,
  author       = {Xiang Zhuang and
                  Qiang Zhang and
                  Keyan Ding and
                  Yatao Bian and
                  Xiao Wang and
                  Jingsong Lv and
                  Hongyang Chen and
                  Huajun Chen},
  title        = {Learning Invariant Molecular Representation in Latent Discrete Space},
  booktitle    = {Proceedings of the 37th Annual Conference
                  on Neural Information Processing Systems (NeurIPS 2023)},
month = {December 10 - 16, 2023},
address = {New Orleans, LA, USA},
  year         = {2023},
  bibsource    = {dblp computer science bibliography, https://dblp.org}
}

@inproceedings{DBLP:conf/nips/ChenBZXHC23,
  author       = {Yongqiang Chen and
                  Yatao Bian and
                  Kaiwen Zhou and
                  Binghui Xie and
                  Bo Han and
                  James Cheng},
  title        = {Does Invariant Graph Learning via Environment Augmentation Learn Invariance?},
  booktitle    = {Proceedings of the 36th Annual Conference on Neural Information Processing Systems (NeurIPS 2023)},
month = {December 10 - 16, 2023},
address= {New Orleans,LA, USA},
  year         = {2023},
  bibsource    = {dblp computer science bibliography, https://dblp.org}
}

@inproceedings{DBLP:conf/nips/WuC0ZL24,
  author       = {Shirley Wu and
                  Kaidi Cao and
                  Bruno Ribeiro and
                  James Y. Zou and
                  Jure Leskovec},
  title        = {GraphMETRO: Mitigating Complex Graph Distribution Shifts via Mixture
                  of Aligned Experts},
  booktitle    = {Proceedings of the 38th Annual Conference
                  on Neural Information Processing Systems 2024 (NeurIPS 2024)},
month = {December 10 - 15, 2024},
address = {Vancouver, BC, Canada},
  year         = {2024},
  bibsource    = {dblp computer science bibliography, https://dblp.org}
}

@inproceedings{DBLP:conf/www/FangLSGZWW024,
  author       = {Junfeng Fang and
                  Xinglin Li and
                  Yongduo Sui and
                  Yuan Gao and
                  Guibin Zhang and
                  Kun Wang and
                  Xiang Wang and
                  Xiangnan He},
  title        = {{EXGC:} Bridging Efficiency and Explainability in Graph Condensation},
  booktitle    = {Proceedings of the 33rd {ACM} on Web Conference 2024 (WWW 2024)},
address= {Singapore},
month = {May 13-17, 2024},
  pages        = {721--732},
  publisher    = {{ACM}},
  year         = {2024},
  doi          = {10.1145/3589334.3645551},
  bibsource    = {dblp computer science bibliography, https://dblp.org}
}

@inproceedings{DBLP:conf/icml/LiWZW0C22,
  author       = {Sihang Li and
                  Xiang Wang and
                  An Zhang and
                  Yingxin Wu and
                  Xiangnan He and
                  Tat{-}Seng Chua},
  title        = {Let Invariant Rationale Discovery Inspire Graph Contrastive Learning},
  booktitle    = {Proceedings of the 34th International Conference on Machine Learning 2022 ({ICML} 2022) },
address = {Baltimore, Maryland, {USA}},
month = {17-23 July, 2022},
  series       = {Proceedings of Machine Learning Research},
  volume       = {162},
  pages        = {13052--13065},
  publisher    = {{PMLR}},
  year         = {2022},
  bibsource    = {dblp computer science bibliography, https://dblp.org}
}

@inproceedings{DBLP:conf/www/GuoZYHW0C21,
  author       = {Zhichun Guo and
                  Chuxu Zhang and
                  Wenhao Yu and
                  John Herr and
                  Olaf Wiest and
                  Meng Jiang and
                  Nitesh V. Chawla},
  title        = {Few-Shot Graph Learning for Molecular Property Prediction},
  booktitle    = {Proceedings of the 30th on The Web Conference 2021 (WWW 2021)},
address ={Ljubljana, Slovenia},
month = {April 19-23, 2021},
  pages        = {2559--2567},
  publisher    = {{ACM} / {IW3C2}},
  year         = {2021},
  url          = {https://doi.org/10.1145/3442381.3450112},
  doi          = {10.1145/3442381.3450112},
  bibsource    = {dblp computer science bibliography, https://dblp.org}
}

@article{doi:10.1021/acscentsci.6b00367,
  author       = {Han Altae{-}Tran and
                  Bharath Ramsundar and
                  Aneesh S. Pappu and
                  Vijay S. Pande},
  title        = {Low Data Drug Discovery with One-shot Learning},
  volume       = {arXiv preprint:1611.03199},
  year         = {2016},
  eprinttype    = {arXiv},
  eprint       = {1611.03199},
  bibsource    = {dblp computer science bibliography, https://dblp.org}
}

@inproceedings{DBLP:conf/cvpr/KimKKY19,
  author       = {Jongmin Kim and
                  Taesup Kim and
                  Sungwoong Kim and
                  Chang D. Yoo},
  title        = {Edge-Labeling Graph Neural Network for Few-Shot Learning},
  booktitle    = {Proceedings of the {IEEE} Conference on Computer Vision and Pattern Recognition 2019 ({CVPR} 2019)  },
address=  {Long Beach, CA, USA},
month = {June 16-20, 2019},
  pages        = {11--20},
  publisher    = {Computer Vision Foundation / {IEEE}},
  year         = {2019},
  bibsource    = {dblp computer science bibliography, https://dblp.org}
}

@inproceedings{DBLP:conf/nips/WangAYD21,
  author       = {Yaqing Wang and
                  Abulikemu Abuduweili and
                  Quanming Yao and
                  Dejing Dou},
  title        = {Property-Aware Relation Networks for Few-Shot Molecular Property Prediction},
  booktitle    = {Proceedings of the 35th Annual Conference
                  on Neural Information Processing Systems 2021 (NeurIPS 2021)},
month = {December 6-14, 2021},
address = {Virtual Event},
pages        = {17441--17454},
  year         = {2021},
  bibsource    = {dblp computer science bibliography, https://dblp.org}
}

@article{DBLP:journals/nn/JuLQFWG0023,
  author       = {Wei Ju and
                  Zequn Liu and
                  Yifang Qin and
                  Bin Feng and
                  Chen Wang and
                  Zhihui Guo and
                  Xiao Luo and
                  Ming Zhang},
  title        = {Few-shot Molecular Property Prediction via Hierarchically Structured
                  Learning on Relation Graphs},
  journal      = {Neural Networks},
  volume       = {163},
  pages        = {122--131},
  year         = {2023},
  bibsource    = {dblp computer science bibliography, https://dblp.org}
}

@inproceedings{DBLP:conf/ijcai/ZhuangZWDFC23,
  author       = {Xiang Zhuang and
                  Qiang Zhang and
                  Bin Wu and
                  Keyan Ding and
                  Yin Fang and
                  Huajun Chen},
  title        = {Graph Sampling-based Meta-Learning for Molecular Property Prediction},
  booktitle    = {Proceedings of the 32nd International Joint Conference on
                  Artificial Intelligence 2023 ({IJCAI} 2023)},
month = {19th-25th August 2023},
address = {Macao,SAR, China},
  pages        = {4729--4737},
  publisher    = {ijcai.org},
  year         = {2023},
  bibsource    = {dblp computer science bibliography, https://dblp.org}
}

@inproceedings{DBLP:conf/sdm/MengLZ0K23,
  author       = {Ziqiao Meng and
                  Yaoman Li and
                  Peilin Zhao and
                  Yang Yu and
                  Irwin King},
  title        = {Meta-Learning with Motif-based Task Augmentation for Few-Shot Molecular
                  Property Prediction},
  booktitle    = {Proceedings of the International Conference on Data Mining 2023 ({ICDM} 2023)},
month = {April 27-29,                  2023},
address = { Paul Twin Cities, MN, USA},
  pages        = {811--819},
  publisher    = {{SIAM}},
  year         = {2023},
  bibsource    = {dblp computer science bibliography, https://dblp.org}
}

@inproceedings{DBLP:conf/icml/FinnAL17,
  author       = {Chelsea Finn and
                  Pieter Abbeel and
                  Sergey Levine},
  title        = {Model-Agnostic Meta-Learning for Fast Adaptation of Deep Networks},
  booktitle    = {Proceedings of the 34th International Conference on Machine Learning 2017 ({ICML} 2017)},
month = {6-11 August 2017},
address ={Sydney, NSW, Australia},
  series       = {Proceedings of Machine Learning Research},
  volume       = {70},
  pages        = {1126--1135},
  publisher    = {{PMLR}},
  year         = {2017},
  bibsource    = {dblp computer science bibliography, https://dblp.org}
}

@inproceedings{DBLP:conf/nips/VinyalsBLKW16,
  author       = {Oriol Vinyals and
                  Charles Blundell and
                  Tim Lillicrap and
                  Koray Kavukcuoglu and
                  Daan Wierstra},
  title        = {Matching Networks for One Shot Learning},
  booktitle    = {Proceedings of the 30th Annual Conference
                  on Neural Information Processing Systems 2016 (NeuIPS 2016)},
address= {Barcelona, Spain},
month = {December 5-10, 2016},
  pages        = {3630--3638},
  year         = {2016},
  bibsource    = {dblp computer science bibliography, https://dblp.org}
}

@inproceedings{DBLP:conf/nips/0006LSW024,
  author       = {Qiang Liu and
                  Shaozhen Liu and
                  Xin Sun and
                  Shu Wu and
                  Liang Wang},
  title        = {Pin-Tuning: Parameter-Efficient In-Context Tuning for Few-Shot Molecular
                  Property Prediction},
  booktitle    = {Proceedings of the 38th on Annual Conference on Neural Information Processing Systems 2024 (NeurIPS 2024)},
  year         = {2024},
month = {December 10 - 15, 2024},
address = {Vancouver,
                  BC, Canada},
  bibsource    = {dblp computer science bibliography, https://dblp.org}
}

@inproceedings{DBLP:conf/nips/SnellSZ17,
  author       = {Jake Snell and
                  Kevin Swersky and
                  Richard S. Zemel},
  title        = {Prototypical Networks for Few-shot Learning},
  booktitle    = {Proceedings of the 31st Annual Conference on Neural Information Processing Systems 2017 (NeuIPS 2017)},
month = {December 4-9, 201},
address = {Long Beach, CA, {USA}},
  pages        = {4077--4087},
  year         = {2017},
  url          = {https://proceedings.neurips.cc/paper/2017/hash/cb8da6767461f2812ae4290eac7cbc42-Abstract.html},
  bibsource    = {dblp computer science bibliography, https://dblp.org}
}

@inproceedings{DBLP:conf/ijcai/MaBALLZL20,
  author       = {Yuqing Ma and
                  Shihao Bai and
                  Shan An and
                  Wei Liu and
                  Aishan Liu and
                  Xiantong Zhen and
                  Xianglong Liu},
  title        = {Transductive Relation-Propagation Network for Few-shot Learning},
  booktitle    = {Proceedings of the 29th International Joint Conference on
                  Artificial Intelligence 2020 ({IJCAI} 2020)},
  pages        = {804--810},
  publisher    = {ijcai.org},
  year         = {2020},
  doi          = {10.24963/IJCAI.2020/112},
  bibsource    = {dblp computer science bibliography, https://dblp.org}
}

@inproceedings{DBLP:conf/icml/AbbasXCCC22,
  author       = {Momin Abbas and
                  Quan Xiao and
                  Lisha Chen and
                  Pin{-}Yu Chen and
                  Tianyi Chen},
  title        = {Sharp-MAML: Sharpness-Aware Model-Agnostic Meta Learning},
  booktitle    = {Proceedings of the 39th International Conference on Machine Learning ({ICML} 2022) },
month = {17-23 July 2022},
address= {Baltimore, Maryland, {USA}},
  volume       = {162},
  pages        = {10--32},
  publisher    = {{PMLR}},
  year         = {2022},
  bibsource    = {dblp computer science bibliography, https://dblp.org}
}

@article{jumper2021highly,
  title={Highly accurate protein structure prediction with AlphaFold},
  author={Jumper, John and Evans, Richard and Pritzel, Alexander and Green, Tim and Figurnov, Michael and Ronneberger, Olaf and Tunyasuvunakool, Kathryn and Bates, Russ and {\v{Z}}{\'\i}dek, Augustin and Potapenko, Anna and others},
  journal={Nature},
  volume={596},
  number={7873},
  pages={583--589},
  year={2021},
  publisher={Nature Publishing Group UK London}
}

@article{DBLP:journals/jcisd/ShenSHCKYWWZZZCCCLZJCJW24,
  author       = {Chao Shen and
                  Jianfei Song and
                  Chang{-}Yu Hsieh and
                  Dong{-}Sheng Cao and
                  Yu Kang and
                  Wenling Ye and
                  Zhenxing Wu and
                  Jike Wang and
                  Odin Zhang and
                  Xujun Zhang and
                  Hao Zeng and
                  Heng Cai and
                  Yu Chen and
                  Linkang Chen and
                  Hao Luo and
                  Xinda Zhao and
                  Tianye Jian and
                  Tong Chen and
                  Dejun Jiang and
                  Mingyang Wang and
                  Qing Ye and
                  Jialu Wu and
                  Hongyan Du and
                  Hui Shi and
                  Yafeng Deng and
                  Tingjun Hou},
  title        = {DrugFlow: An AI-Driven One-Stop Platform for Innovative Drug Discovery},
  journal      = {J. Chem. Inf. Model.},
  volume       = {64},
  number       = {14},
  pages        = {5381--5391},
  year         = {2024},
  bibsource    = {dblp computer science bibliography, https://dblp.org}
}

@article{DBLP:journals/jcisd/RusinkoRFBKG24,
  author       = {Andrew Rusinko III and
                  Mohammad A. Rezaei and
                  Lukas Friedrich and
                  Hans{-}Peter Buchstaller and
                  Daniel Kuhn and
                  Ashwini Ghogare},
  title        = {{AIDDISON:} Empowering Drug Discovery with {AI/ML} and {CADD} Tools
                  in a Secure, Web-Based SaaS Platform},
  journal      = {J. Chem. Inf. Model.},
  volume       = {64},
  number       = {1},
  pages        = {3--8},
  year         = {2024},
  doi          = {10.1021/ACS.JCIM.3C01016},
  bibsource    = {dblp computer science bibliography, https://dblp.org}
}

@article{liu2020drugcombdb,
  title={DrugCombDB: a comprehensive database of drug combinations toward the discovery of combinatorial therapy},
  author={Liu, Hui and Zhang, Wenhao and Zou, Bo and Wang, Jinxian and Deng, Yuanyuan and Deng, Lei},
  journal={Nucleic acids research},
  volume={48},
  number={D1},
  year={2020},
  publisher={Oxford University Press},
}

\newpage
~
\newpage

\section{Appendix}

\subsection{A. Theoretical Analysis}
\label{app:theory_1}

%\resizebox{0.44\textwidth}{!}{$    $}

We provide a theoretical analysis of CaMol through an information-theoretic lens, e.g., mutual information, to explain how CaMol discovers causal substructures while isolating them from noisy ones.
Given an input molecule $G$, we assume that it contains a causal substructure $C$ that is sufficient for determining its chemical property $Y$. 
Formally, this implies that the prediction depends only on the causal component $C$ within $G$, i.e., $p(Y \mid G) = p(Y \mid C, G)$, where $p$ denotes the conditional distribution of the property label $Y$.
We now approximate the true distribution $p$ with a predictive model $q$ from CaMol.
To learn this approximation, we minimize the negative conditional log-likelihood of the observed labels.  
The objective function is defined as:
\begin{align}
-\mathcal{L} = - \sum_{i=1}^{n} \log q(Y_i \mid C_i, G_i),
\end{align}
where $q(Y_i \mid C_i, G_i)$ indicates the probability that the model assigns to the label $Y_i$ given the molecule $G_i$ and its causal substructure $C_i$.
We now decompose this loss as follows:  
\begin{equation}
\label{eq:mi_1}
\resizebox{0.43\textwidth}{!}{$ 
\begin{aligned}
- \mathcal{L} &= - \sum_{i=1}^n \log \left[ \frac{q(Y_i \mid C_i)}{p(Y_i \mid C_i)} 
\cdot \frac{p(Y_i \mid C_i)}{p(Y_i \mid G_i)}  
\cdot p(Y_i \mid G_i) \right]  \\
&= - \sum_{i=1}^n \left[ \log \frac{q(Y_i \mid C_i)}{p(Y_i \mid C_i)} 
+ \log \frac{p(Y_i \mid C_i)}{p(Y_i \mid G_i)} 
+ \log p(Y_i \mid G_i) \right] \\
&= \mathbb{E}  \left[ \log \frac{q(Y_i \mid C_i)}{p(Y_i \mid C_i)} \right]
+ \mathbb{E}  \left[ \log \frac{p(Y_i \mid C_i)}{p(Y_i \mid G_i)} \right] 
- \mathbb{E}  \left[ \log p(Y_i \mid G_i)  \right] 
\end{aligned}
$}
\end{equation}

Then, the second term can be represented as:

\begin{equation}
\label{eq:mi_2}
\resizebox{0.43\textwidth}{!}{$ 
\begin{aligned}
\mathbb{E} \left[ \log \frac{p(Y_i \mid C_i)}{p(Y_i \mid G_i)} \right]
&= \mathbb{E} \left[ \log \frac{p(Y_i \mid C_i)}{p(Y_i \mid C_i, S_i)} \right]   \\
&=  \sum_i p(G_i, Y_i) \log \frac{p(Y_i \mid C_i)}{p(Y_i \mid C_i, S_i)}  \\
&= \sum_i p(G_i, Y_i) \log \left( 
\frac{p(Y_i \mid C_i)}{p(Y_i \mid C_i, S_i)} 
\cdot \frac{p(S_i \mid C_i)}{p(S_i \mid C_i)}
\right)  \\
&=  \sum_i p(G_i, Y_i) \log \frac{p(S_i, Y_i \mid C_i)}{p(Y_i \mid C_i) \cdot p(S_i \mid C_i)} \\
&=  I(S; Y \mid C)
\end{aligned}
$}
\end{equation}

By plugging Equation \ref{eq:mi_2} into Equation \ref{eq:mi_1}, we obtain the final objective function, as:
\begin{equation}
\label{eq:mi_3}
\begin{aligned}
\min \ \mathbb{E} \left[ \log \frac{q(Y_i \mid C_i)}{p(Y_i \mid C_i)} \right]
+ I(S; Y \mid C)
+ H(Y \mid G),
\end{aligned}
\end{equation}
where the first term denotes the KL divergence between the model distribution $q$ and the true distribution $p$ given causal substructure $C$, 
the second term captures the spurious dependence between noise $S$ and labels $Y$ conditioned on $C$,
and the third term denotes the entropy of labels $Y$ given the whole molecule $G$.
%Here, our objective is to focus on the second term, which quantifies the conditional dependency between the label $Y$ and the confounding substructure $S$ given the known causal substructure $C$.
Here, our primary interest lies in the second term $I(S; Y \mid C)$, which measures how much information the confounding substructures $S$ still provide about the labels once the causal substructure $C$ is known.
By applying the chain rule, we have:
\begin{equation}
\label{eq:mi_4}
\begin{aligned}
I\left(S; Y \mid C\right) =  I\left (S; Y, C \right) - I\left(S; C\right).
\end{aligned}
\end{equation}
This implies that minimizing the second term in Equation \ref{eq:mi_3} encourages the causal substructure $C$ to capture sufficient information about $Y$.
We interpret the behavior of CaMol from two complementary perspectives:

\noindent \textbf{(1) Identifying causal substructures.}
$I(S; Y \mid C)$ measures the residual dependence between noisy substructures $S$ and labels $Y$ given the causal substructures $C$.
Minimizing this term ensures that $C$ captures all the information about the label $Y$, isolating $S$ as redundant.
That is, CaMol could disentangle $C$ from $S$, isolating $C$ as the true substructures that are strongly correlated to $Y$.

 \noindent \textbf{(2) Removing confounding substructures.}
The residual dependence $I(S; Y)$ quantifies the spurious correlation between noisy substructures $S$ and the target property $Y$.
By minimizing this dependency, CaMol suppresses non-causal signals and prevents the model from treating $S$ as informative for predicting $Y$, effectively cutting off the direct path $S \rightarrow R \rightarrow Y$ (Figure~\ref{fig:Few_shot}).
This ensures that the effect of $C$ on $Y$ is estimated faithfully, aligning predictions with invariant causal mechanisms rather than specific biases.

% The residual dependence $I(S; Y)$ quantifies the noisy correlation between $S$ and $Y$.
% By minimizing this dependency, CaMol suppresses the noise and prevents the model from treating $S$ as informative for $Y$, as we are solely interested in cutting down the direct path between $S \rightarrow R  \rightarrow Y$ (Figure \ref{fig:Few_shot}).
% This ensures that the effect of $C$ on $Y$ is estimated faithfully, aligning predictions with invariant causal mechanisms rather than specific biases.
%\end{itemize}

\begin{algorithm}[h]
\DontPrintSemicolon
\KwIn{Meta-learning model parameters $\theta$, inner learning rate $\alpha_{\text{inner}}$, meta learning rate $\alpha_{\text{outer}}$, number of tasks per epoch $B$.}
\KwOut{Optimized meta-parameter $\theta$}

%\For{epoch $= 1$ to $N_{\text{epochs}}$}{
\While{not converge}{
    $\mathcal{L}_{\text{total}} \leftarrow 0$ \;

    \For{step $= 1$ to $B$}{
        Sample $B$ episode from training set $\mathcal{D}_{\text{train}}$ \;
        
        Compute support loss $\mathcal{L}_S(f_\theta)$ on $S$ by Eq. \ref{eq:inner-loss}  \;
    
        Compute gradients: $\nabla_{\theta} \mathcal{L}_S(f_\theta)$  \;
        
        Update: $\theta' \leftarrow \theta - \alpha_{\text{inner}} \nabla_{\theta_i} \mathcal{L}_S(f_\theta)$ by Eq. \ref{eq:inner-update} \;
         Compute loss $\mathcal{L}_Q(f_{\theta'})$ on query set $Q$  by Eq. \ref{eq:outer-loss} \;
    
        $\mathcal{L}_{\text{total}} \leftarrow \mathcal{L}_{\text{total}} + \mathcal{L}_Q(f_{\theta'})$ \;
    }
    $\mathcal{L}_{\text{total}} \leftarrow \mathcal{L}_{\text{total}} / B$ \;
    Update meta-parameters: $\theta \leftarrow \theta - \alpha_{\text{outer}} \nabla_\theta \mathcal{L}_{\text{total}}$ 
}
Return optimized parameter $\theta$.
\caption{Training process of CaMol}
\label{alg:fs}
\end{algorithm}

\subsection{B. Training Process of CaMol}
\label{app:alg}

The training process of CaMol follows a meta-learning paradigm based on episodic optimization, as shown in Algorithm \ref{alg:fs}.
%In each meta-training epoch, a batch of few-shot tasks is sampled, where each task is split into a support set and a query set.
For every task, CaMol initializes task-specific parameters from the shared meta-parameters and performs several inner-loop updates using the support set, guided by the causal substructure generator and distribution intervention.
The adapted parameters are then evaluated on the query set, and the resulting query losses are aggregated across tasks to compute the meta-loss.
Finally, in the outer-loop optimization, the meta-loss is used to update the global parameters.

\subsection{C. Statistics of Datasets}
\label{appendix:dataset}

\begin{table}[t]
\centering
\caption{An overview of statistics of datasets.}
\setlength{\tabcolsep}{ 1 mm }% Default value: 6pt
\begin{adjustbox}{width= \linewidth}
\begin{tabular}{l ccc ccc}
\toprule
{Dataset} & {Tox21} & {SIDER} & {MUV} & {ToxCast} & {PCBA} & ClinTox \\
\midrule
\# Compound & 7,831 & 1,427 & 93,127 & 8,575 & 437,929 &  1,478 \\
\# Property & 12 & 27 & 17 & 617 & 128& 2 \\
\# Train Property & 9 & 21 & 12 & 451 & 118 & 1 \\
\# Test Property & 3 & 6 & 5 & 158 & 10 & 1 \\
\midrule
\% Label active & 6.24 & 56.76 & 0.31 & 12.60 & 0.84  & 50.61\\
\% Label inactive & 76.71 & 43.24 & 15.76 & 72.43 & 59.84 & 49.39\\
\% Missing Label & 17.05 & 0.00 & 84.21 & 14.97 & 39.32  & 0.00\\
\bottomrule
\end{tabular}
\label{tab:dataset_stats}
\end{adjustbox}
\end{table}

We evaluate CaMol on six widely used molecular property prediction benchmarks from MoleculeNet, namely Tox21, SIDER, MUV, ToxCast, PCBA, and ClinTox \cite{wu2018moleculenet}, as shown in Table~\ref{tab:dataset_stats}.
We follow the standard data splits commonly adopted in Few-shot MPP to ensure fair comparison with prior work \cite{doi:10.1021/acscentsci.6b00367}.
In addition, to evaluate the interpretability of CaMol, we further evaluate on three datasets with ground-truth substructure annotations: Benzene, Alkane Carbonyl, and Fluoride Carbonyl \cite{agarwal2023evaluating}.

\begin{figure*}[tb]
\centering
\begin{subfigure}{\linewidth}
    \centering
    \includegraphics[width=\linewidth]{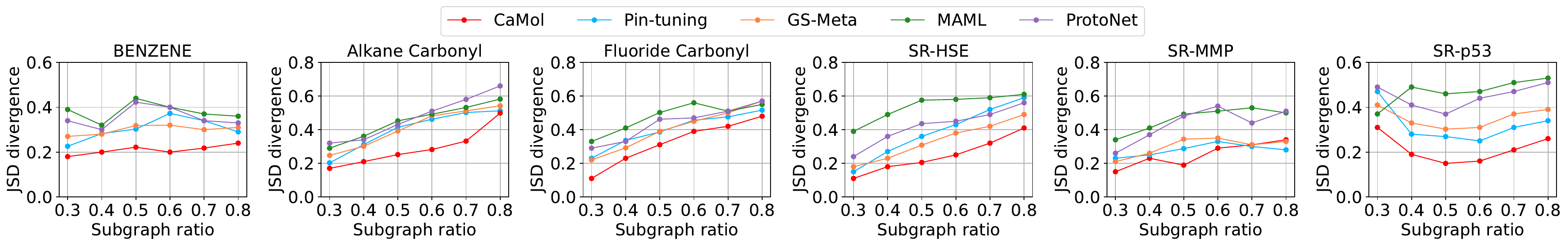}
    \caption{A comparison on consistency of representations of causal substructures according to their sizes (subgraph ratios) in terms of JSD.
    %
    %JSD scores for found causal substructure embeddings with different subgraph ratios.
    }
    \label{fig:sen_camol_interpretrability_6}
\end{subfigure}
%\vspace{0.8em} % optional spacing between figures
\begin{subfigure}{\linewidth}
    \centering
    \includegraphics[width=\linewidth]{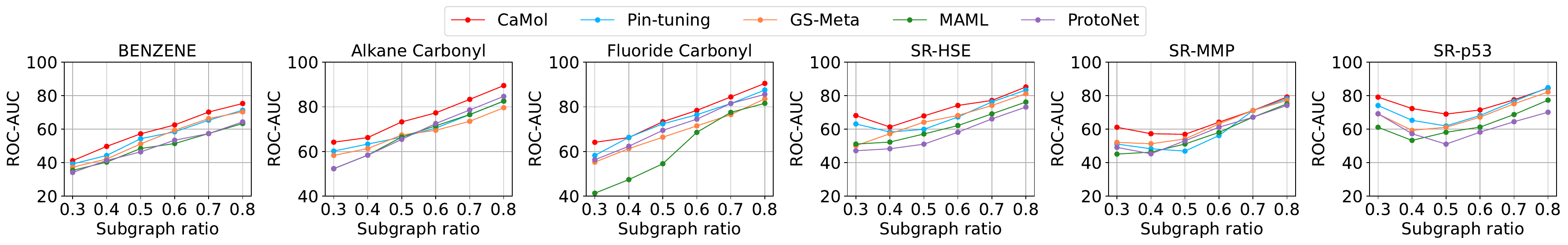}
    \caption{A performance comparison on Few-shot MPP according to sizes of discovered causal substructures (subgraph ratios) in terms of ROC-AUC.
    }
    \label{fig:sen_camol_acc}
\end{subfigure}

\caption{Sensitivity comparisons on subgraph ratios in terms of (a) consistency and (b) few-shot performance. Models should discover consistent causal substructures for a property, contributing to prediction accuracy, regardless of substructures sizes.}
\label{fig:sen_camol_subgraph}
\end{figure*}

\subsection{D. Implementation Details}
\label{app:imp_details}

\subsubsection{D.1. Model Hyperparameters}

The detailed hyperparameters of CaMol are provided in Table~\ref{tab:appendix_hyperparameters}.
We train the model with the Adam optimizer, using an initial learning rate of $1\times 10^{-4}$, weight decay of $1\times 10^{-5}$, and a batch size of 8 for 3000 epochs.
The context graph encoder is implemented as a 3-layer GIN with edge feature learning \cite{DBLP:conf/iclr/XuHLJ19}, using an input feature dimension of 64.
For graph-level readout, we apply a summation pooling function to aggregate node representations.
For meta-learning, the inner-loop and outer-loop learning rates are set to 0.05 and 0.001, respectively.

\subsubsection{D.2. Baselines}

We compare CaMol against two groups of few-shot baselines, including traditional methods and in-context learning methods.
For traditional methods, we adopt five approaches:  
\begin{itemize}
    \item \textbf{MAML}~\cite{DBLP:conf/icml/FinnAL17} is an optimization-based meta-learning framework that learns a shared initialization across tasks.  
    
    \item \textbf{Sharp-MAML}~\cite{DBLP:conf/icml/AbbasXCCC22} extends MAML by incorporating sharpness-aware minimization during training to better generalize in few-shot adaptation.

    \item \textbf{ProtoNet}~\cite{DBLP:conf/nips/SnellSZ17} is a metric-based method that represents each class by a prototype in the latent space.
    %by measuring distances to these prototypes.

    \item \textbf{EGNN}~\cite{DBLP:conf/cvpr/KimKKY19} uses edge-aware GNNs to capture relational information about output representations in the latent space.
    
    \item \textbf{Meta-MGNN}~\cite{DBLP:conf/www/GuoZYHW0C21} uses a meta-learning framework with self-supervised task adaptation to enable effective few-shot MPP tasks.
\end{itemize}

For in-context learning methods, we adopt five recent methods:

\begin{itemize}
    \item \textbf{PAR}~\cite{DBLP:conf/nips/WangAYD21} presents prototype-based adaptation with relation networks between molecules to improve few-shot MPP. 
    
    \item \textbf{GS-Meta}~\cite{DBLP:conf/ijcai/ZhuangZWDFC23} models inter-task dependencies through exploiting relationships between molecules.
    
    \item \textbf{TPN}~\cite{DBLP:conf/ijcai/MaBALLZL20} uses a context graph to perform label propagation between molecules and properties.

    \item \textbf{HSL-RG}~\cite{DBLP:journals/nn/JuLQFWG0023} employs hierarchical structure learning with relational graphs to capture multi-level contextual information to benefit in-context few-shot adaptation.

    \item \textbf{Pin-Tuning}~\cite{DBLP:conf/nips/0006LSW024} constructs a context graph of molecule and property relationships.  
    By propagating labels through the context graph under learnable adapter layers, it improves adaptability to unseen molecular properties.
\end{itemize}

\subsubsection{D.3. Training Resources}

All experiments were run on two servers, each equipped with four NVIDIA RTX A5000 GPUs (24 GB memory per GPU).
Our model was implemented in Python 3.8.8, using the PyTorch Geometric framework \cite{FeyLenssen2019} and the DGL \cite{wang2019dgl}.
Experiments were conducted in an Ubuntu 20.04 LTS environment.

\subsubsection{D.4. Reproducibility}
To ensure reproducibility, we release an anonymized implementation of our model, including the full source code and experimental setups, in an open repository\footnote{\url{https://anonymous.4open.science/r/CaMol-18A7}}.

% We provide the source code for our proposed model. 
% An anonymized implementation containing the code and experiment setups is available at the repository\footnote{\url{https://github.com/anonymoussubmissionwww2026-prog/CaMol}}.

% To ensure reproducibility, we provide detailed instructions for training and evaluating our proposed model. 
% An anonymized implementation, including experiment configurations and scripts, is available at the repository\footnote{\url{https://github.com/anonymoussubmissionwww2026-prog/CaMol}}.  

\begin{table}[tb]
\caption{Hyperparameters of CaMol used in experiments.}
\centering
\setlength{\tabcolsep}{16 pt}% Default value: 6pt
%\fontsize{11 pt}{13 pt}\selectfont
%\small
%\begin{adjustbox}{width= \linewidth}
\begin{tabular}{l c }
\toprule
Hyperparameters
&Values
\\ 
\midrule

Number of EGIN layers 
&3
\\
Initial feature dimension
&64
\\
Number of training epochs
&3000
\\ 
Adam: initial learning rate
&$1\times 10^{-4}$
\\
Adam: weight decay
& $1\times 10^{-5}$
\\
% Activation function
% & ReLU
% \\
Readout function
& SUM
\\
Hyperparameter $\alpha_1$
&0.1
\\
Hyperparameter $\alpha_2$
&0.01
\\
Learning rate in inner-loop $\alpha_{\text{inner}}$
&0.05
\\
Learning rate in outer-loop $\alpha_{\text{outer}}$
&0.001
\\
\bottomrule
\end{tabular}
%\end{adjustbox}
\label{tab:appendix_hyperparameters}
\end{table}

\subsection{E. Additional Experiments}

\begin{figure}[h]
\centering
  \includegraphics[width=  \linewidth]{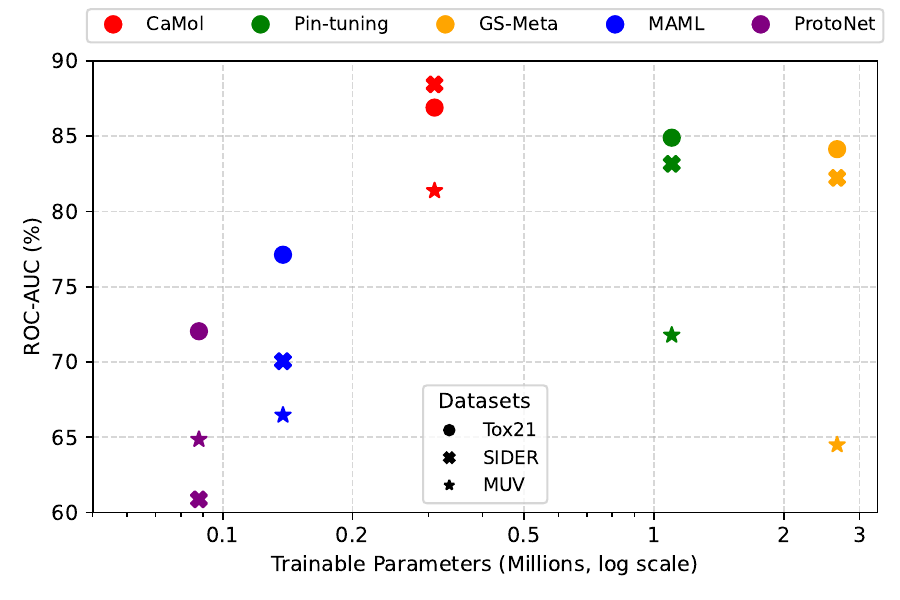}
  \caption{A parameter efficiency comparison on few-shot MPP in terms of ROC-AUC and trainable parameter sizes.
  %Trade-off between model accuracy and trainable parameter size on Tox21, SIDER, and MUV datasets.
  }
  \label{fig:Trade-off}
\end{figure}

\begin{figure}[h]
\centering
  \includegraphics[width=  \linewidth]{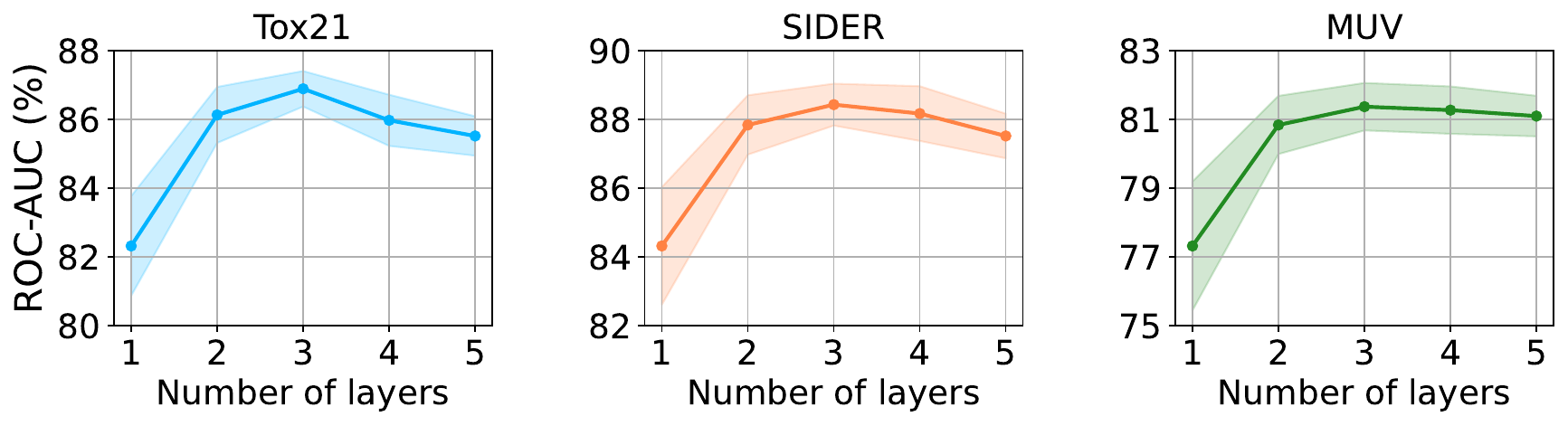}
  \caption{Performance with varying numbers of GNN layers.}
  \label{fig:Sen_GNNlayers}
\end{figure}

\begin{figure}[!ht]
\centering
\begin{subfigure}{\linewidth}
    \centering
    \includegraphics[width=\linewidth]{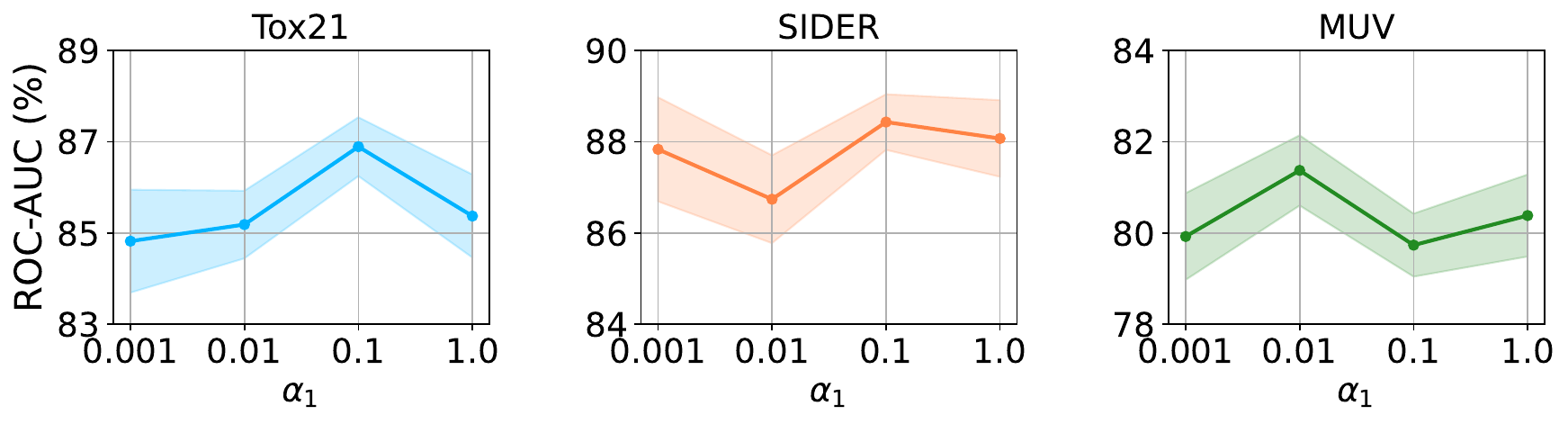}
    \caption{Sensitivity analysis on hyperparameter $\alpha_1$.}
    \label{fig:Sen_alpha_1}
\end{subfigure}

\vspace{0.5em} % optional space between the two subfigures

\begin{subfigure}{\linewidth}
    \centering
    \includegraphics[width=\linewidth]{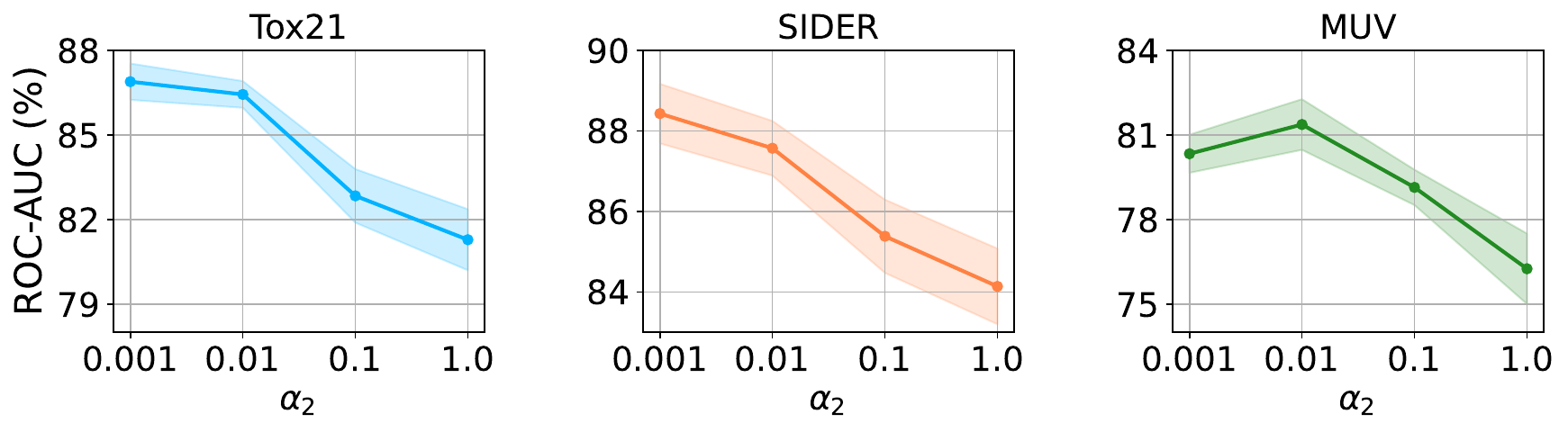}
    \caption{Sensitivity analysis on hyperparameter $\alpha_2$.}
    \label{fig:Sen_alpha_2}
\end{subfigure}
\caption{Sensitivity analysis of the weights in loss function.}
\label{fig:sensitivity_alpha}
\end{figure}

\subsubsection{E.1. Efficiency Analysis of Model Performance versus Parameter Size}
\label{app:exp_trade_off}

Figure~\ref{fig:Trade-off} presents the trade-off between accuracy and parameter size on Tox21, SIDER, and MUV.
We observed that CaMol achieved the best accuracy efficiency balance: with only 0.31M parameters, it consistently outperformed in-context methods, e.g., Pin-tuning (1.10M) and GS-Meta (2.66M).
This implies that causal substructure discovery provides a stronger inductive bias than simply scaling model size and normal adaptation strategies.
In contrast, models with more parameters do not guarantee better accuracy, and traditional meta-learning methods, while parameter-efficient, underperformed significantly.
Overall, CaMol lies on the Pareto frontier, combining high accuracy with compact parameters.

\subsubsection{E.2. Sensitivity Analysis on the Subgraph Ratios}
\label{app:exp_sub_rarios}

% We further conduct a sensitivity analysis by varying the causal subgraph ratios, ranging from 0.3 to 0.8, and evaluating the JSD score of the found causal substructure embeddings, as shown in Figure \ref{fig:sen_camol_interpretrability_6}.
% We observed that CaMol achieved the lowest JSD scores across all ratios.
% This indicates that the found subgraphs remain consistent with the specific properties.
% Moreover, the mid-range ratios, i.e., around 0.5–0.6, obtain a good balance between compactness and sufficiency.
% That is, very small ratios could discard important atoms and increase divergence, while very large ratios reintroduce noisy context that dilutes causal signals.
% This indicates CaMol’s robustness in identifying minimal yet consistent causal substructures, thereby enhancing both molecule interpretability and few-shot performance.
We conducted a sensitivity analysis by varying the causal subgraph ratio from 0.3 to 0.8 and evaluating the JSD of the found causal substructures, as shown in Figure~\ref{fig:sen_camol_interpretrability_6}.
We observed that CaMol consistently achieved the lowest JSD across all ratios, indicating robustness in capturing stable and consistent substructures.
The mid-range ratios, i.e., 0.5–0.7, show the best trade-off between sufficiency and performance, indicating that the small ratios attempt to discard important atoms, while very large ratios could introduce noisy signals.
These results demonstrate CaMol’s ability to extract minimal yet consistent causal substructures, improving both interpretability and few-shot predictive performance.
We further evaluated predictive performance across six benchmarks, as shown in Figure~\ref{fig:sen_camol_acc}.
%CaMol achieved the best results, improving with larger subgraph ratios and outperforming baselines, e.g., Pin-tuning and GS-Meta.
Unlike baselines that degrade sharply under small ratios, from 0.3 to 0.5, CaMol maintained competitive accuracy with only marginal losses, demonstrating its robustness in leveraging limited information while scaling effectively with sufficient contexts.
% We further evaluate molecular property prediction across six benchmark datasets in terms of ROC-AUC, as shown in Figure \ref{fig:sen_camol_acc}.
% We observed that CaMol’s performance steadily increases with larger subgraph ratios and consistently achieves the highest ROC-AUC compared to baselines such as Pin-tuning, GS-Meta, MAML, and ProtoNet.
% This indicates that retaining more nodes could help our proposed model enhance predictive accuracy.
% Moreover, while baselines exhibit significant degradation under small ratios (e.g., 0.3–0.5), CaMol maintains competitive performance with only marginal losses.
% These results demonstrate that CaMol effectively leverages limited subgraph information while scaling gracefully with larger contexts, providing a stable and reliable solution for molecular property prediction.
%CaMol consistently achieves the highest ROC-AUC, demonstrating both superior accuracy and robustness compared to other baselines, such as Pin-tuning, GS-Meta, MAML, and ProtoNet.
%conduct a sensitivity analysis by varying the subgraph ratio from 0.3 to 0.8 and evaluate molecular property prediction across six datasets, as shown in Figure \ref{fig:sen_camol_acc}.

\subsubsection{E.3. Sensitivity Analysis of the Weights for Loss Terms} 
\label{app:sen_1}

We analyze the sensitivity of CaMol to the hyperparameters $\alpha_{1}$ and $\alpha_{2}$, which control the weights of $\mathcal{L}_{\text{KL}}$ and $\mathcal{L}_{\text{var}}$, respectively, as shown in Figure~\ref{fig:sensitivity_alpha}.
For $\alpha_{1}$, CaMol remains stable across a wide range and achieves peak performance at moderate values, e.g., 86.89 ROC-AUC on Tox21.
%, 88.43 on SIDER. 
The small values underweight interventions, while overly large ones slightly degrade accuracy.
For $\alpha_{2}$, performance is strongest at small values, e.g., $0.001$–$0.01$, but drops at larger ones, e.g., $0.1$–$1.0$, indicating that semantic confounders are helpful but need to be balanced with the context signals.

\subsubsection{E.4. Sensitivity Analysis on the Number of Layers} 
\label{app:sen_2}

We investigated the effect of varying the number of GNN layers on CaMol’s performance across three datasets: Tox21, SIDER, and MUV, as shown in Figure~\ref{fig:Sen_GNNlayers}.
We observed that the model performance is improved as depth increases from 1 to 3 layers, with peak performance observed at 2 or 3 layers, e.g., 86.89 on Tox21 and 88.43 on SIDER.
Moreover, additional layers yield marginal or negative gains, suggesting that deeper models introduce over-smoothing without capturing additional useful structure.

\end{document}